%% file: ms.tex
\documentclass[twocolumn,10pt]{article}
\usepackage[
top=1.78cm,
bottom=1.78cm,
left=1.4cm,
right=1.4cm,
headsep=0cm,
]{geometry}
\usepackage{etoolbox}
\usepackage[T1]{fontenc}
\usepackage{amsmath,amssymb,amsfonts}
\usepackage{algorithmic}
\usepackage{graphicx}
\usepackage{placeins}
\usepackage{setspace}
\usepackage{enumitem}
\usepackage{titlesec}
\usepackage{textcomp}
\usepackage[dvipsnames]{xcolor}
\usepackage{siunitx}
\usepackage[export]{adjustbox}
\usepackage{blindtext}
\usepackage{booktabs}
\usepackage{tikz}
\usepackage{multirow}
\usepackage{mwe,tikz}\usepackage[percent]{overpic}
\usepackage[font=footnotesize]{subcaption}
\usepackage[font=small, skip=5pt]{caption}
\usepackage[affil-it]{authblk}
\usepackage[style=ieee]{biblatex}
\usetikzlibrary{shapes}
\usetikzlibrary{shapes.misc}
\tikzset{cross/.style={cross out, draw=black, minimum size=2*(#1-\pgflinewidth), inner sep=0pt, outer sep=0pt},
cross/.default={0.6mm}}
\DeclareSymbolFont{matha}{OML}{txmi}{m}{it}% txfonts
\DeclareMathSymbol{\varv}{\mathord}{matha}{118}
\def\BibTeX{{\rm B\kern-.05em{\sc i\kern-.025em b}\kern-.08em
    T\kern-.1667em\lower.7ex\hbox{E}\kern-.125emX}}
    
\renewrobustcmd{\bfseries}{\fontseries{b}\selectfont}
\renewrobustcmd{\boldmath}{}
\newrobustcmd{\B}{\bfseries}

\AtBeginEnvironment{tabular}{\footnotesize}
\AtBeginEnvironment{abstract}{\itshape}
\AtBeginBibliography{\fontsize{8.5pt}{9.75pt}\selectfont}
\setlist{nosep}

\DeclareMathOperator*{\argmin}{arg\,min}

\DeclareSIUnit\pixel{px}

\makeatletter
\newcommand*\bigcdot{\mathpalette\bigcdot@{.5}}
\newcommand*\bigcdot@[2]{\mathbin{\vcenter{\hbox{\scalebox{#2}{$\m@th#1\bullet$}}}}}
\makeatother

\renewcommand{\thesection}{\Roman{section}}
\renewcommand{\thesubsection}{\Alph{subsection}} 
\renewcommand{\thesubsubsection}{\arabic{subsubsection}}
\titleformat{\section}{\normalfont\large\scshape\centering}{\thesection.}{0.5em}{}
\titleformat{\subsection}{\normalfont\itshape}{\thesubsection.}{0.5em}{}
\titleformat{\subsubsection}[runin]{\normalfont\itshape}{\thesubsubsection)}{0.5em}{}[:]
\titlespacing{\section}{0pt}{1.5ex plus 0.5ex minus .2ex}{1.0ex plus .2ex}
\titlespacing{\subsection}{0pt}{1.1ex plus 0.5ex minus .2ex}{0.5ex plus .2ex}
\titlespacing{\subsubsection}{\parindent}{0pt}{0.5em}% left-before-after
\makeatletter
\renewcommand{\p@section}{}
\renewcommand{\p@subsection}{\thesection-}
\renewcommand{\p@subsubsection}{\thesection-\thesubsection.}
\makeatother

\definecolor{light-gray}{gray}{0.95}
\AtBeginEnvironment{snugshade*}{\vspace{-\FrameSep}}
\AfterEndEnvironment{snugshade*}{\vspace{-\FrameSep}}

\makeatletter
\def\footnoterule{\relax%
  \kern-5pt
  \hbox to \columnwidth{\vrule width 0.45\columnwidth height 0.4pt\hfill}
  \kern4.6pt
  }
\makeatother

\makeatletter 
\newcommand\semiHuge{\@setfontsize\semiHuge{22.72}{29.5}}
\makeatother
\makeatletter
\patchcmd{\@maketitle}{\@title}{\vspace{-1cm}\semiHuge \@title}{}{}
\patchcmd{\@maketitle}{\@author}{\normalsize\@author}{}{}
\makeatother
\begin{document}
\title{Georeferencing of Photovoltaic Modules from Aerial Infrared Videos using Structure-from-Motion}  % and its Application to Thermal Analysis of PV Plants

\author[1]{Lukas Bommes}
\author[1]{Claudia Buerhop-Lutz}
\author[1]{Tobias Pickel}
\author[1]{Jens Hauch}
\author[1,2]{Christoph Brabec}
\author[1]{Ian Marius Peters}
\affil[1]{Forschungszentrum Jülich GmbH, Helmholtz-Institute Erlangen-Nuremberg for Renewable Energies (HI ERN)}
\affil[2]{Institute Materials for Electronics and Energy Technology, Universität Erlangen-Nürnberg (FAU)\par\normalsize\normalfont{Correspondence to i.peters@fz-juelich.de}}

\date{}

\maketitle
\thispagestyle{empty} % prevent page number on first page

\begin{abstract}
\begin{spacing}{0.9}
\input{abstract}
\end{spacing}
\end{abstract}

% \begin{IEEEkeywords}
%         PV Plant Inspection, PV Module Detection, Structure-from-motion, 3D Reconstruction, Geospatial, Mapping, Drone, Thermography, Large-Scale Dataset, Instance Segmentation
% \end{IEEEkeywords}

\input{introduction}
\input{method}
\input{results}
\input{conclusion}
\input{acknowledgements}

\FloatBarrier

%\bibliography{ms.bib}
%\bibliographystyle{ieeetr}
\printbibliography

\newpage
\onecolumn

\appendix
\input{appendix}

\end{document}

%% file: abstract.tex
To identify abnormal photovoltaic (PV) modules in large-scale PV plants economically, drone-mounted infrared (IR) cameras and automated video processing algorithms are frequently used. While most related works focus on the detection of abnormal modules, little has been done to automatically localize those modules within the plant. In this work, we use incremental structure-from-motion to automatically obtain geocoordinates of all PV modules in a plant based on visual cues and the measured GPS trajectory of the drone. In addition, we extract multiple IR images of each PV module. Using our method, we successfully map \SI{99.3}{\percent} of the $35084$ modules in four large-scale and one rooftop plant and extract over $2.2$ million module images. As compared to our previous work, extraction misses $18$ times less modules (one in $140$ modules as compared to one in eight). Furthermore, two or three plant rows can be processed simultaneously, increasing module throughput and reducing flight duration by a factor of $2.1$ and $3.7$, respectively. Comparison with an accurate orthophoto of one of the large-scale plants yields a root mean square error of the estimated module geocoordinates of \SI{5.87}{\meter} and a relative error within each plant row of \SI{0.22}{\meter} to \SI{0.82}{\meter}. Finally, we use the module geocoordinates and extracted IR images to visualize distributions of module temperatures and anomaly predictions of a deep learning classifier on a map. While the temperature distribution helps to identify disconnected strings, we also find that its detection accuracy for module anomalies reaches, or even exceeds, that of a deep learning classifier for seven out of ten common anomaly types. The software is published at {\color{magenta}https://github.com/LukasBommes/PV-Hawk}.

%% file: introduction.tex
\section{Introduction}

%[intro]
% The large amounts of global installed solar photovoltaics (PV) and expected future growth require automatic image analysis for adequate quality control. As PV modules may develop defects due to environmental influences, aging or incorrect handling, PV plants need to be inspected regularly to ensure safe operation and maximum yield. Due to the large size of most PV plants, inspection is only economic if highly automated \cite{Bizzarri.2019}. Thus, recent years have seen a surge in automated PV plant inspection systems \cite{Bommes.2021, Henry.2020, Carletti.2019, Grimaccia.2017, Francesco.2018, Niccolai.2019, Arenella.2017, Addabbo.2018, Deitsch.2016, Dunderdale.2020}. These systems rely on drones equipped with a thermal infrared (IR) camera, that enables detection of abnormal PV modules based on their thermal signature \cite{Gallardo-Saavedra.2018, Niccolai.2019b, Kumar.2018, Buerhop.2012, Buerhop.2012b, Scheuerpflug.2014, Quater.2014}. The large amounts of acquired IR images are automatically processed by computer vision algorithms, which detect PV modules in the images \cite{Diaz.2020, Carletti.2019, Greco.2020, Deitsch.2016}, predict module anomalies \cite{Bommes.2021b, Dunderdale.2020, Addabbo.2018, Deitsch.2016}, and localize each module in the PV plant \cite{Grimaccia.2017, Niccolai.2019, Tsanakas.2016, Nisi.2016}.

% one reviewer did not agree with the clustering of references
%[intro]
The large amount of global installed solar photovoltaics (PV) and expected future growth require automatic image analysis for adequate quality control. As PV modules may develop defects due to environmental influences, aging or incorrect handling, PV plants need to be inspected regularly to ensure safe operation and maximum yield. Due to the large size of most PV plants, inspection is only economic if highly automated \cite{Bizzarri.2019}. Thus, recent years have seen a surge in automated PV plant inspection systems, such as the ones by Zefri et al.~\cite{Zefri.2022}, Pierddicca et al.~\cite{Pierdicca.2020}, Henry et al.~\cite{Henry.2020}, and Carletti et al.~\cite{Carletti.2019}. These systems rely on drones equipped with a thermal infrared (IR) camera, that enables detection of abnormal PV modules based on their thermal signature \cite{Gallardo-Saavedra.2018}. The large amounts of acquired IR images are automatically processed by computer vision algorithms, which typically detect PV modules in the images, predict module anomalies, and localize each module in the PV plant.

% these references are removed from the work
% Niccolai.2019b, Kumar.2018, Buerhop.2012, Buerhop.2012b, Scheuerpflug.2014, Quater.2014
% Diaz.2020, Greco.2020, Dunderdale.2020, Arenella.2017, Deitsch.2016

% these references were added to the work
% Pierdicca.2020, Zefri.2022
% Denz.2020, Rajendran.2016, Nomura.2017, Lappalainen.2016, Wan.2021

%[problem definition]
In this work, we focus on the localization of PV modules in large-scale plants. Localization is a crucial task as it enables targeted repairs of abnormal modules. However, it is also notoriously difficult to identify the correct module among millions of identically looking and densely packed modules from a highly repetitive video with only a limited viewport. Previous works attempted to solve this problem by stitching adjacent video frames of a PV plant row into a panorama image \cite{Grimaccia.2017, Francesco.2018, Aghaei.2016}. This approach was successful, yet only works well for short video sequences. And, as also shown in our previous work \cite{Bommes.2021}, panorama stitching requires manual selection of the video frames for each row and provides the module location only relative to other modules. Niccolai et al. \cite{Niccolai.2019} also use panorama stitching and additionally match each row panorama to a CAD plan. While this yields absolute module locations, it requires a CAD plan, which is not always available and, even if it is available, is by no means standardized across different PV plants.

Other works explore direct georeferencing of PV modules in each image based on the measured GPS position and altitude of the drone \cite{Addabbo.2018, Nisi.2016}. Georeferencing requires a centimeter-accurate Realtime Kinematics GPS (RTK-GPS) and is prone to GPS measurement errors as no additional visual cues are considered. Further, georeferencing is limited to nadiral images, which may contain sun reflections and exhibit sub-optimal contrast compared to images taken under the optimal viewing angle. Being limited to nadiral images also makes drone operation more difficult.

Another method for module localization is the creation of an orthophoto from a few high-altitude images \cite{Lee.2019, Zefri.2018, Tsanakas.2016}. Orthophotos allow visualizing the temperature distribution of the entire PV plant. One issue with this approach is that it is not always possible to take images from high altitudes, e.g. if there are nearby streets. Furthermore, a low spatial resolution and possible visual artefacts impede accurate detection of abnormal modules based on the orthophoto alone.

%[what does our method do better than the related works?]
This work presents a new method for PV plant inspection based on aerial IR videos. As opposed to the related works, our method is fully automated, provides the absolute geocoordinates of each PV module instead of a relative location, works on long video sequences of large-scale plants, requires no CAD plan, works with both standard GPS and RTK-GPS and is not limited to nadiral videos. Furthermore, videos can be acquired from low flight altitudes and multiple high resolution images of each module are obtained, which are important for downstream analysis.

%[description of our method]
Our method builds on our previous work \cite{Bommes.2021}, but features a more general approach for PV module localization based on structure-from-motion (SfM) \cite{Schoenberger.2016, Agarwal.2009} to obtain absolute geocoordinates of the PV modules in a plant. In addition, PV module images are extracted from each video frame and tracked over subsequent frames. Based on the extracted images, module anomalies can be detected with a deep learning classifier \cite{Bommes.2021b} and visualized on a map. This enables quick assessment of the health state of the entire PV plant and helps performing targeted repairs. Similarly, module temperatures can be mapped across the PV plant. Temperature mapping allows detecting abnormal modules by comparison with neighbouring modules. This approach can replace more complex deep learning classifiers for detecting abnormal modules, as we will show. As opposed to the related works, our method relies on both visual cues and measured GPS trajectory for georeferencing. This improves robustness to GPS measurement errors and allows to use standard GPS instead of RTK-GPS. We further use videos instead of individual images. Video analysis speeds up data acquisition and works not only with automatic waypoint flights, but also with manual flights performed ad-hoc for small and irregular plants. Videos also yield larger amounts of data as each PV module is captured in multiple video frames, which is beneficial for training machine learning algorithms on the extracted data. Compared to our previous work, requirements on the flight trajectory are less stringent, and, in principle, plants with non-row layouts, such as rooftop plants, can be processed. Having a single tool for different types of PV plants is more cost-effective and requires less maintenance than multiple plant-specific solutions. We also show that for regular plants, multiple rows can be scanned simultaneously, which significantly increases throughput.

%% file: method.tex
\section{Method}
\label{sec:method}

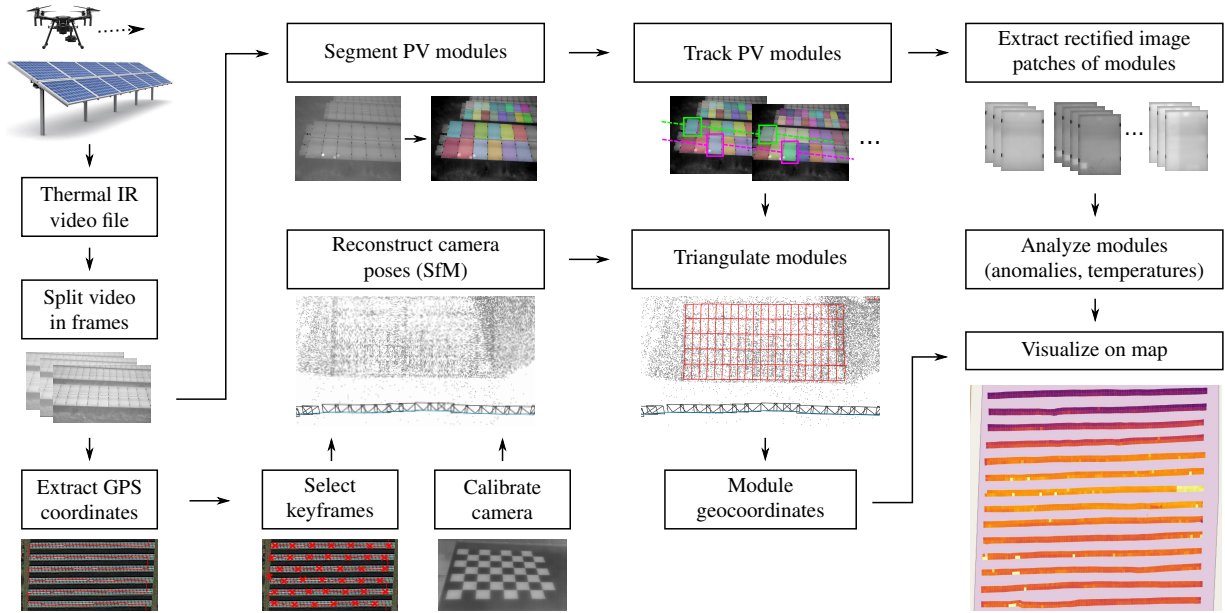
\begin{figure*}[tbp]
    \centering
    \begin{overpic}[scale=1.0, abs, unit=1mm, tics=5]{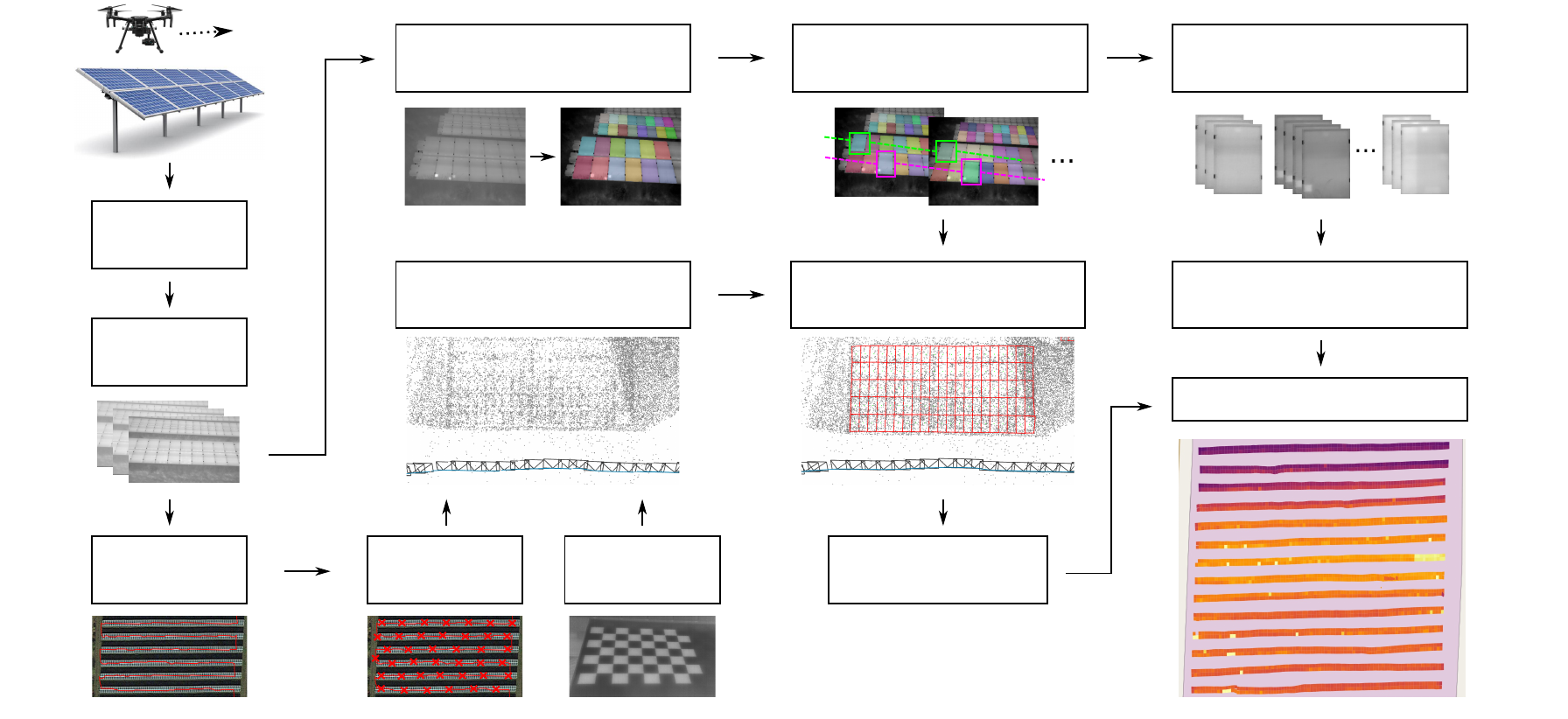}
    \put (10.583, 60.239) {\parbox[t][8mm][c]{18.205mm}{\footnotesize \centering Thermal IR\\video file}}
    \put (10.583, 46.601) {\parbox[t][8mm][c]{18.205mm}{\footnotesize \centering Split video\\in frames}}
    \put (10.583, 21.314) {\parbox[t][8mm][c]{18.205mm}{\footnotesize \centering Extract GPS\\coordinates}}
    \put (42.563, 21.314) {\parbox[t][8mm][c]{18.205mm}{\footnotesize \centering Select\\keyframes}}
    \put (65.482, 21.314) {\parbox[t][8mm][c]{18.205mm}{\footnotesize \centering Calibrate\\camera}}
    \put (45.872, 80.751) {\parbox[t][8mm][c]{34.363mm}{\footnotesize \centering Segment PV modules}}
    \put (91.997, 80.751) {\parbox[t][8mm][c]{34.363mm}{\footnotesize \centering Track PV modules}}
    \put (45.872, 53.262) {\parbox[t][8mm][c]{34.363mm}{\footnotesize \centering Reconstruct camera\\poses (SfM)}}
    \put (91.708, 53.262) {\parbox[t][8mm][c]{34.363mm}{\footnotesize \centering Triangulate modules}}
    \put (96.074, 21.314) {\parbox[t][8mm][c]{25.631mm}{\footnotesize \centering Module\\geocoordinates}}
    \put (136.087, 80.751) {\parbox[t][8mm][c]{34.363mm}{\footnotesize \centering Extract rectified image\\patches of modules}}
    \put (136.087, 53.262) {\parbox[t][8mm][c]{34.363mm}{\footnotesize \centering Analyze modules\\(anomalies, temperatures)}}
    \put (136.087, 39.697) {\parbox[t][5.2mm][c]{34.363mm}{\footnotesize \centering Visualize on map}}
    %\put (58.058, 70.89) {\parbox[t][10.642mm][c]{10.332mm}{\footnotesize \centering Mask\\R-CNN}}
    %\put (46.3, 58.5) {\footnotesize Frame}
    %\put (73, 58.5) {\footnotesize Masks}
    %\put (140, 69) {\footnotesize 2bc}
    %\put (149.3, 69) {\footnotesize fa3}
    %\put (161, 69) {\footnotesize 1cc}
    %\put (96, 68.2) {\footnotesize \color{white} 2bc}
    %\put (95, 63) {\footnotesize \color{white} fa3}
    %\put (105, 67.4) {\footnotesize \color{white} 2bc}
    %\put (104.5, 61) {\footnotesize \color{white} fa3}
    %\put (115.5, 66) {\footnotesize \color{white} 2bc}
    %\put (117, 60.5) {\footnotesize \color{white} fa3}
    %\put (98, 9.5) {\footnotesize 2bc}
    %\put (103.5, 7.5) {\footnotesize fa3}
    %\put (111.5, 8.5) {\footnotesize 1cc}
    \end{overpic}
    \caption{Overview of our method for automatic extraction and georeferencing of PV modules from aerial IR videos.}
    \label{fig:method_overview}
\end{figure*}

This section introduces our method for fully automatic extraction and georeferencing of PV modules from aerial IR videos. For an overview see fig.~\ref{fig:method_overview}. After acquisition with the drone, IR videos of a PV plant are split into individual frames and the GPS trajectory of the drone is extracted and interpolated. Following Bommes et al.~\cite{Bommes.2021}, PV modules are segmented by Mask R\nobreakdash-CNN \cite{He.2017}, tracked over subsequent frames, extracted and stored to disk. To georeference PV modules, a subset of keyframes is selected based on travelled GPS distance and visual overlap. Subsequently, a georeferenced $3$D reconstruction of the PV plant is obtained by incremental SfM alongside the $6$-DOF camera pose of each keyframe. This requires calibrated camera parameters, which are obtained beforehand. The known keyframe poses are then used to triangulate observed PV modules into the $3$D reconstruction, yielding the desired module geocoordinates.

% how to get x coordinate and width/height of parbox: select surrounding box in Inkscape, read X, W, H attributes
% how to compute y cooridnate: need to invert y axis, so compute Y' = image_height - Y (here: image_height=83.5mm)

%###########################################################################################

\subsection{Camera Model and Calibration}
\label{sec:camera_model_and_calibration}

Several steps of our pipeline use a calibrated pinhole camera model to project $3$D scene points into image coordinates and to triangulate image points into a $3$D reconstruction of the scene. Lens distortion is modelled by a Brown-Conrady radial distortion model \cite{Brown.1966} with five distortion coefficients. 

\begin{figure}[tbp]
    \centering
    \includegraphics[width=0.5\linewidth]{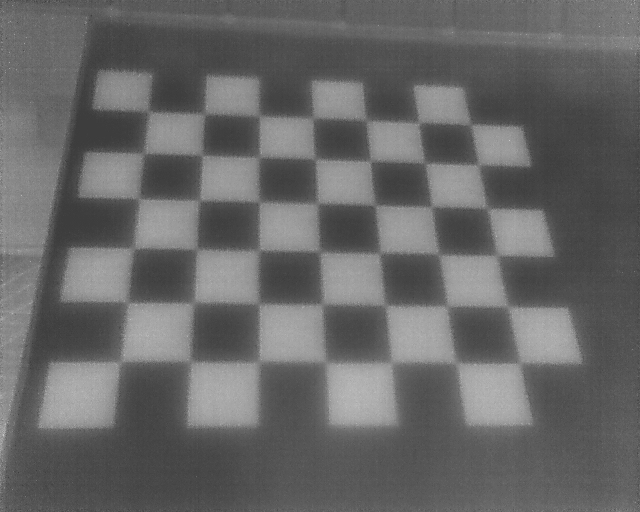}
    \caption{Examplary IR image of the camera calibration target.}
    \label{fig:calibration_target}
\end{figure}

Calibration is performed once for each camera using OpenCV's \cite{Bradski.2000} calibration method with around $150$ images of a chessboard calibration target (see fig.~\ref{fig:calibration_target}). The target consists of foil patches applied to a polymer panel, providing sufficient contrast in the IR image due to different emissivities. We obtain best results when capturing calibration images outside on a cloudy day.

%###########################################################################################

\subsection{Drone Flight and Video Acquisition}

Our method is intended to be used with IR videos acquired by a drone, which scans one or multiple rows of a PV plant at an altitude of \SI{10}{\meter} to \SI{30}{\meter} and at a velocity that ensures blur-free images. Acquisition should take place under clearsky conditions and solar irradiance above \SI{700}{\watt\per\square\meter}. Similar to our previous method \cite{Bommes.2021}, both nadiral and non-nadiral videos can be processed and the camera angle and flight velocity may be varied during the flight. For accurate georeferencing of the SfM reconstruction the drone needs to travel a sufficient distance in at least two orthogonal directions. Furthermore, the flight altitude should be kept approximately constant in case standard GPS is used and no accurate altitude measurement is available. These requirements are much less restrictive than those of our previous work, resulting in higher flexibility and robustness. It is, for example, no problem, if the drone moves non-monotonically along a plant row, or if the same row is scanned multiple times. Furthermore, situations, in which the scanned row is cropped at the top or bottom of the frame, or in which other rows become visible in the camera viewport, can be handled.

For compatibility with the remaining processing steps, we split the acquired IR videos into individual $16$\nobreakdash-bit grayscale images, convert each image to Celsius scale, normalize to the interval $\left[0, 255 \right]$ using its minimum and maximum temperature value, convert to $8$\nobreakdash-bit, and finally, perform histogram equalization.

%###########################################################################################

\subsection{Segmentation, Tracking and Extraction of PV modules}

These steps correspond to our previous work \cite{Bommes.2021} and are therefore described only briefly. A Mask R\nobreakdash-CNN instance segmentation model, which is trained on a photovoltaic-specific dataset, obtains a binary segmentation mask for each PV module in each video frame. After fitting a quadrilateral to each mask, the underlying image region is extracted, warped to a rectangular region by a homography and stored as a $16$\nobreakdash-bit radiometric image file. A tracking algorithm associates masks of the same PV module over subsequent frames and assigns a unique tracking ID to each module. The tracking ID is then used to group the extracted image patches of each module.

%###########################################################################################

\subsection{Preprocessing of the GPS Trajectory}

The drone records its latitude and longitude in WGS\nobreakdash-$84$ coordinates at a rate of \SI{1}{\hertz}. As we do not use RTK-GPS, the measured altitude is unreliable and we assume it as unknown in the subsequent steps. To match the rates of GPS measurements to the higher frame rate of the camera, we perform piecewise linear interpolation of the GPS trajectory and sample a GPS position for each frame. Prior to this, we transform the trajectory from WGS\nobreakdash-$84$ coordinates to local tangent plane (LTP) coordinates \cite{Torge.2012}. LTP coordinates are Cartesian with their origin at or near the inspected site. This enables accurate interpolation and enhances numerical stability in the subsequent SfM procedure.

%###########################################################################################

\subsection{Selection of Frames for Reconstruction}

In this step we select a subset of partially overlapping video frames for the subsequent SfM procedure, which we call keyframes. Subsampling the frames keeps the computational cost of the SfM procedure, which is quadratic in the number of frames, within an acceptable range. It also decouples SfM from the video frame rate, simplifying the use of different cameras. SfM further benefits from the larger parallax between any two keyframes, ensuring more accurate triangulation of scene points. 

We select a frame as a keyframe if i) its distance to the previous keyframe along the GPS trajectory exceeds \SI{0.75}{\meter}, or if ii) its intersection over union (IoU) with the previous keyframe is smaller than \SI{85}{\percent}. To obtain the IoU, ORB features \cite{Rublee.2011} of the frame and the previous keyframe are extracted and matched. A homography is estimated from the matches, which projects the bounding rectangle of the frame onto that of the previous keyframe. The IoU is then the intersection area of both rectangles divided by their total area.

One advantage of capturing videos over individual images, is the ability to adjust the overlap between images after the data is already captured.

%###########################################################################################

\subsection{Reconstruction of Camera Poses with SfM}

In this step the $6$-DOF camera pose of each keyframe is reconstructed using OpenSfM, an incremental SfM library \cite{OpenSfM}. Inputs are the calibrated camera parameters and the selected keyframes with their GPS positions in LTP coordinates. Due to unavailability of reliable measurements we set the GPS altitude to zero and fix the dilution of precision (DOP) to \SI{0.1}{\meter}. Outputs are the rotation and translation of each keyframe in a LTP coordinate system and a $3$D point cloud of reconstructed scene points, which is not further needed. An example is shown in fig.~\ref{fig:reconstruction_map_points_angled}. In the following, we explain briefly how the SfM library works.

\subsubsection{Feature detection and matching}

The SfM library first finds HAHOG features \cite{Mikolajczyk.2002}, i.e. characteristic points, in each keyframe. Overlapping frames are then found by matching these features between pairs of frames. To limit the search space matches are computed only for frame pairs which are at most \SI{15}{\meter} apart.

\subsubsection{Initialization of the reconstruction}
One frame pair with sufficient parallax is selected for initialization of the reconstruction. The pose of the first frame is set as world coordinate origin. The pose of the second frame relative to the first frame is estimated with the five-point algorithm \cite{Nister.2004} or, in case of a planar scene, by decomposing a homography \cite{Faugeras.1988}. An initial set of $3$D scene points is triangulated from the matched feature points in both frames.

\subsubsection{Iterative reconstruction}
Starting from the initial frame pair the other keyframes are added incrementally to the reconstruction. In each iteration the frame with most matches to any of the reconstructed frames is selected. Its pose is estimated from observed $3$D scene points in the reconstruction and their corresponding $2$D projections in the frame by solving the perspective-n-point problem \cite{Lepetit.2008}. Subsequently, new scene points are triangulated from feature points shared between the newly added frame and other frames in the reconstruction. Afterwards, the entire reconstruction is rigidly transformed, so that camera positions best align with their measured GPS positions. In regular intervals bundle adjustment optimizes all reconstructed camera poses and scene points simultaneously by minimizing the reprojection error of the scene points in all frames. Here, camera positions are kept close to their measured GPS positions. Additionally, camera parameters are refined. We use these refined parameters in all subsequent steps.

\subsubsection{Post-processing of the reconstruction}
Under some circumstances the reconstruction of a long video sequence can fail partially. This results in multiple partial reconstructions each with a different LTP coordinate origin. To register all partial reconstructions in a common LTP coordinate frame, we transform each partial reconstruction to WGS-84 coordinates using the reconstruction-specific LTP origin. We then transform the reconstruction back to LTP coordinates, this time using the common LTP origin. This common origin is arbitrarily set to the origin of the first partial reconstruction.

%###########################################################################################

\subsection{Obtaining Geocoordinates of PV Modules}

\begin{figure*}[tbp]
     \captionsetup[subfigure]{aboveskip=2pt, belowskip=0pt, justification=centering}
     \centering
     \begin{subfigure}[t]{0.37\linewidth}
         \centering
         \includegraphics[width=\linewidth, trim=2cm 0.6cm 0.6cm 3cm, clip]{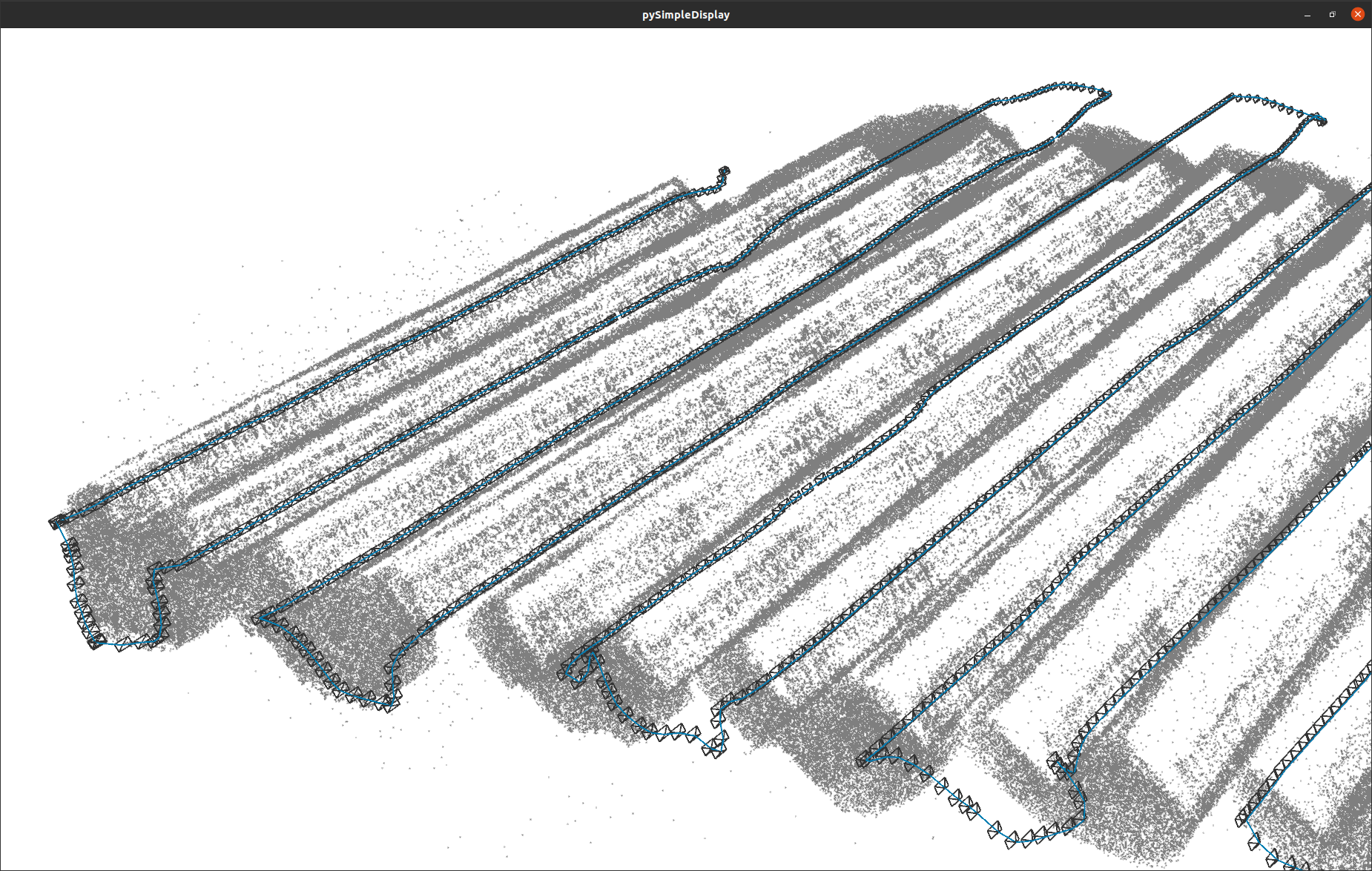}
        \caption{}
        \label{fig:reconstruction_map_points_angled}
     \end{subfigure}
     \hfill
     \begin{subfigure}[t]{0.37\linewidth}
         \centering
         \includegraphics[width=\linewidth, trim=2cm 0.6cm 0.6cm 3cm, clip]{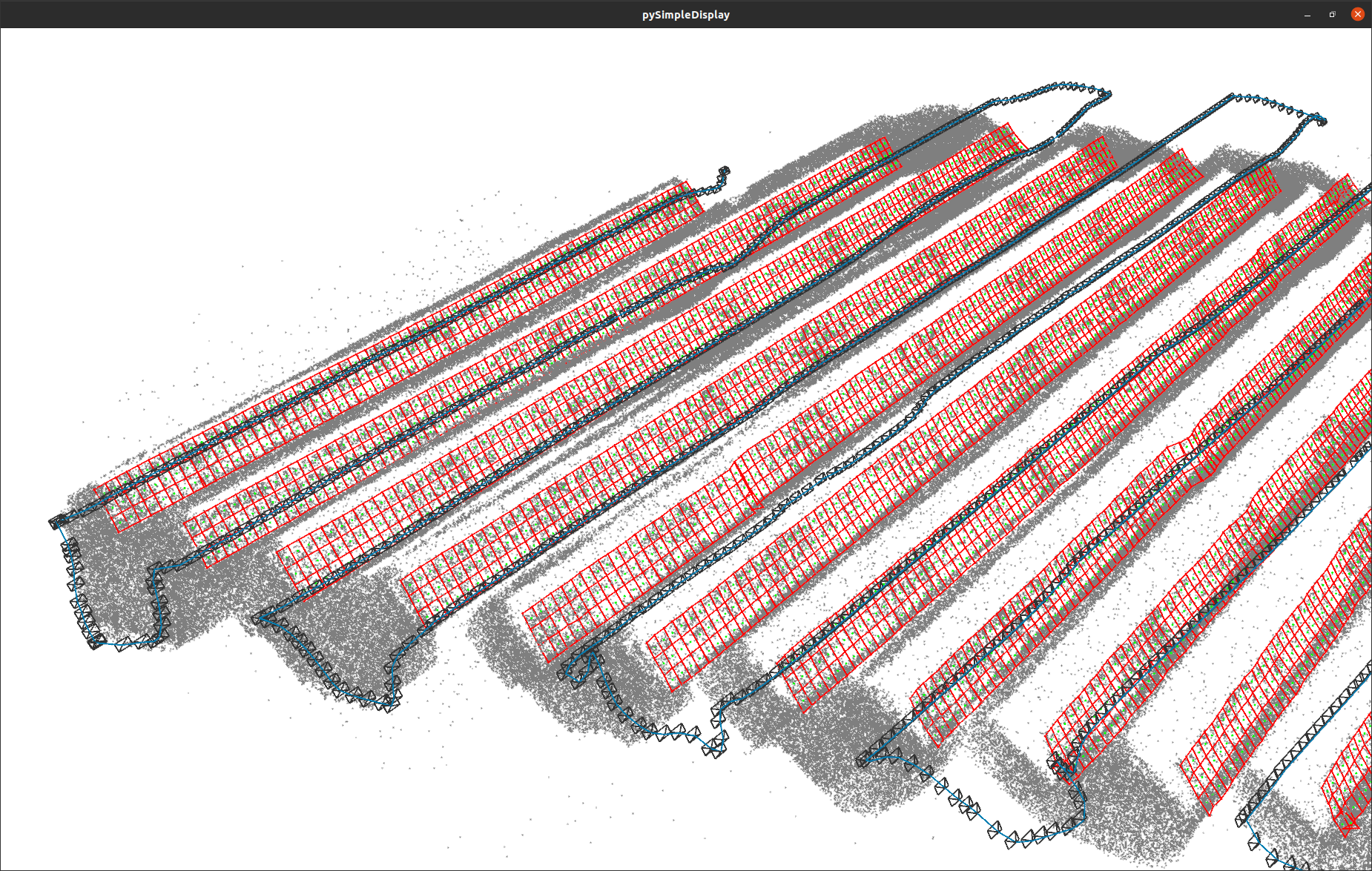}
        \caption{}
        \label{fig:reconstruction_map_points_with_modules_angled}
     \end{subfigure}
     \hfill
     \begin{subfigure}[t]{0.215\linewidth}
         \centering
         \includegraphics[width=\linewidth, trim=11.5cm 0.6cm 15.5cm 1.7cm, clip]{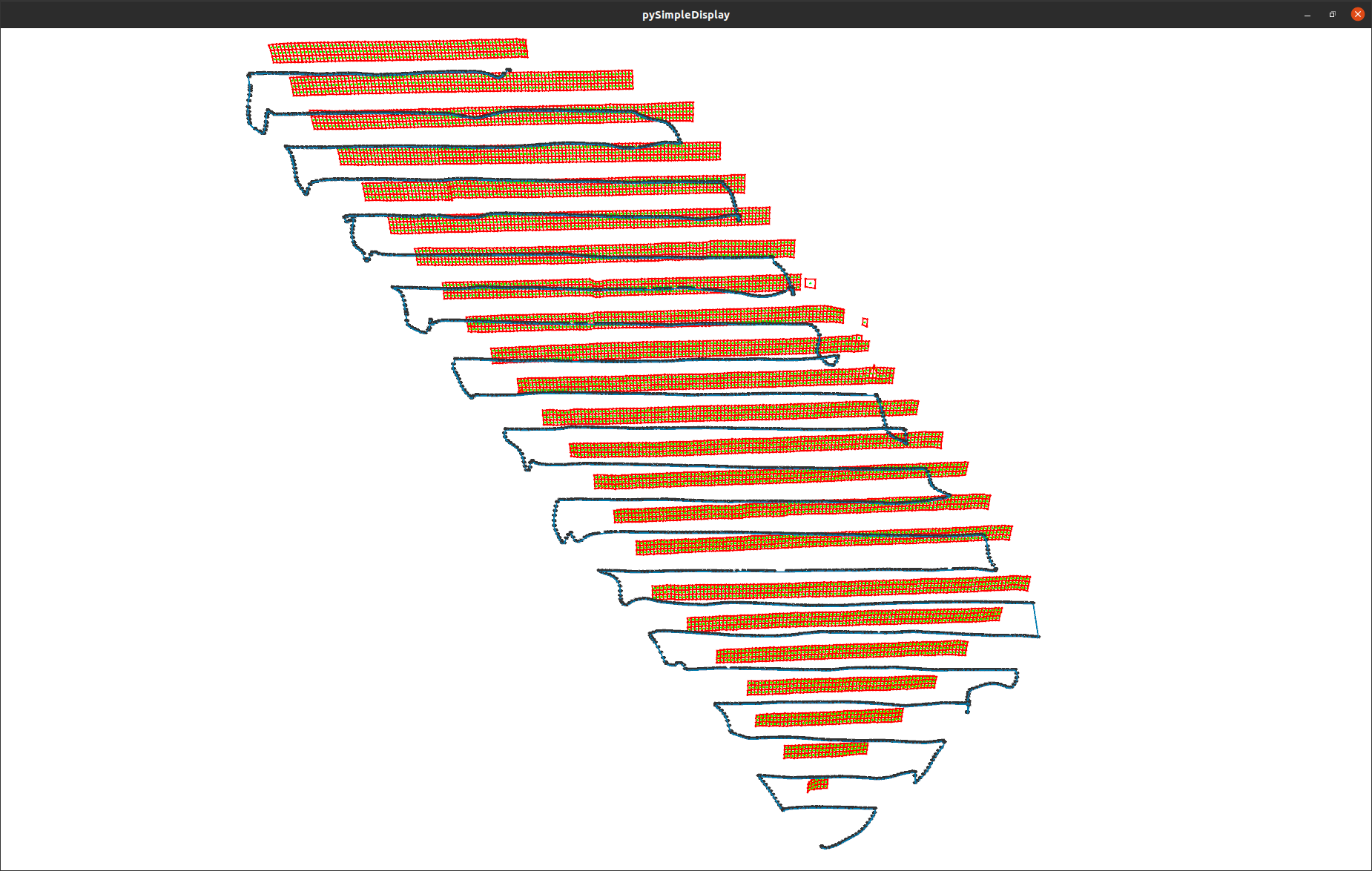}
        \caption{}
        \label{fig:reconstruction_whole_plant_top_down}
     \end{subfigure}
        \caption{Reconstruction results of the SfM procedure: a) reconstructed feature points (grey) and camera poses (blue line and black camera frustrums), b) with triangulated PV modules, c) top-down view on the triangulated modules of an entire PV plant.}
        \label{fig:reconstruction_result}
\end{figure*}

Once the keyframe poses are reconstructed, we triangulate the corner/center points of segmented PV modules into the reconstruction, yielding corresponding LTP geocoordinates. Examples of this are shown in fig.~\ref{fig:reconstruction_map_points_with_modules_angled} and \ref{fig:reconstruction_whole_plant_top_down}. Due to inaccuracies in the module segmentation a robust triangulation procedure and subsequent refinement of the obtained LTP coordinates are required.

\subsubsection{Triangulation of PV modules}

For each tracked module, we obtain pixel-coordinates of the four corner points and the center point in all keyframes, in which the module is visible. Modules observed in less than two keyframes are skipped as they are likely spurious detections and triangulation is impossible. The five module points are then undistorted with the calibrated Brown-Conrady model, and triangulated from all possible pairs of keyframes, in which they are observed. The so triangulated points are only retained if the following two conditions are met: i) The angle between the two viewing rays is larger than \SI{1}{\degree} for all five points, and ii) none of the reprojection errors of the five points exceeds a threshold of $5$ pixels. As there are typically several pairs of keyframes observing the same module, we get a noisy set of triangulated modules (see fig.~\ref{fig:module_triangulation_before_median}). We fuse them robustly by computing the median of corresponding points (see fig.~\ref{fig:module_triangulation_before_merging}).

\subsubsection{Merging of duplicate detections}
\label{sec:merging_of_duplicate_detections}

A PV module may be lost during tracking and reappear a few frames later with a different tracking ID, resulting in multiple overlapping triangulations of the module in the reconstruction (see fig.~\ref{fig:module_triangulation_before_merging}). This step identifies and fuses such duplicates. To this end, for each keyframe, all triangulated modules are projected back into the frame. Two or more modules are identified as overlapping if the mean Euclidean distance between the corresponding four corner points and the center point is smaller than $20$ pixels. To merge overlapping modules, module points are re-triangulated according to the procedure above, this time using all keyframes of the overlapping modules. The result is shown in fig.~\ref{fig:module_triangulation_before_refinement}.

\subsubsection{Refinement of triangulated modules}

\begin{figure}[tbp]
     \captionsetup[subfigure]{aboveskip=2pt, belowskip=0pt, justification=centering}
     \centering
     \begin{subfigure}[t]{0.49\linewidth}
         \centering
         \includegraphics[width=\linewidth, trim=0.3cm 1.5cm 0.5cm 3cm, clip]{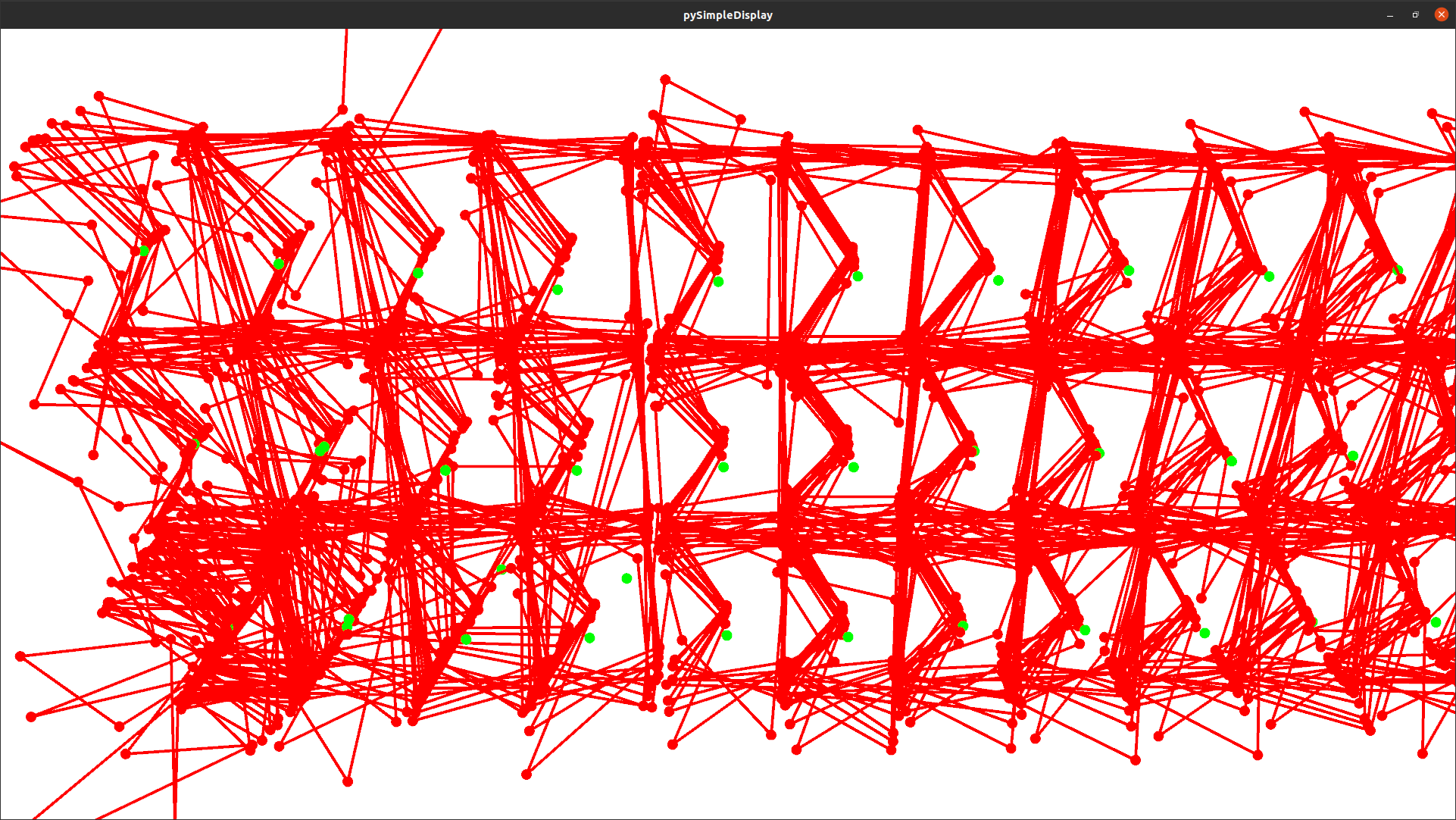}
        \caption{}
        \label{fig:module_triangulation_before_median}
     \end{subfigure}
     \hfill
     \begin{subfigure}[t]{0.49\linewidth}
         \centering
         \includegraphics[width=\linewidth, trim=1cm 3cm 0.5cm 6cm, clip]{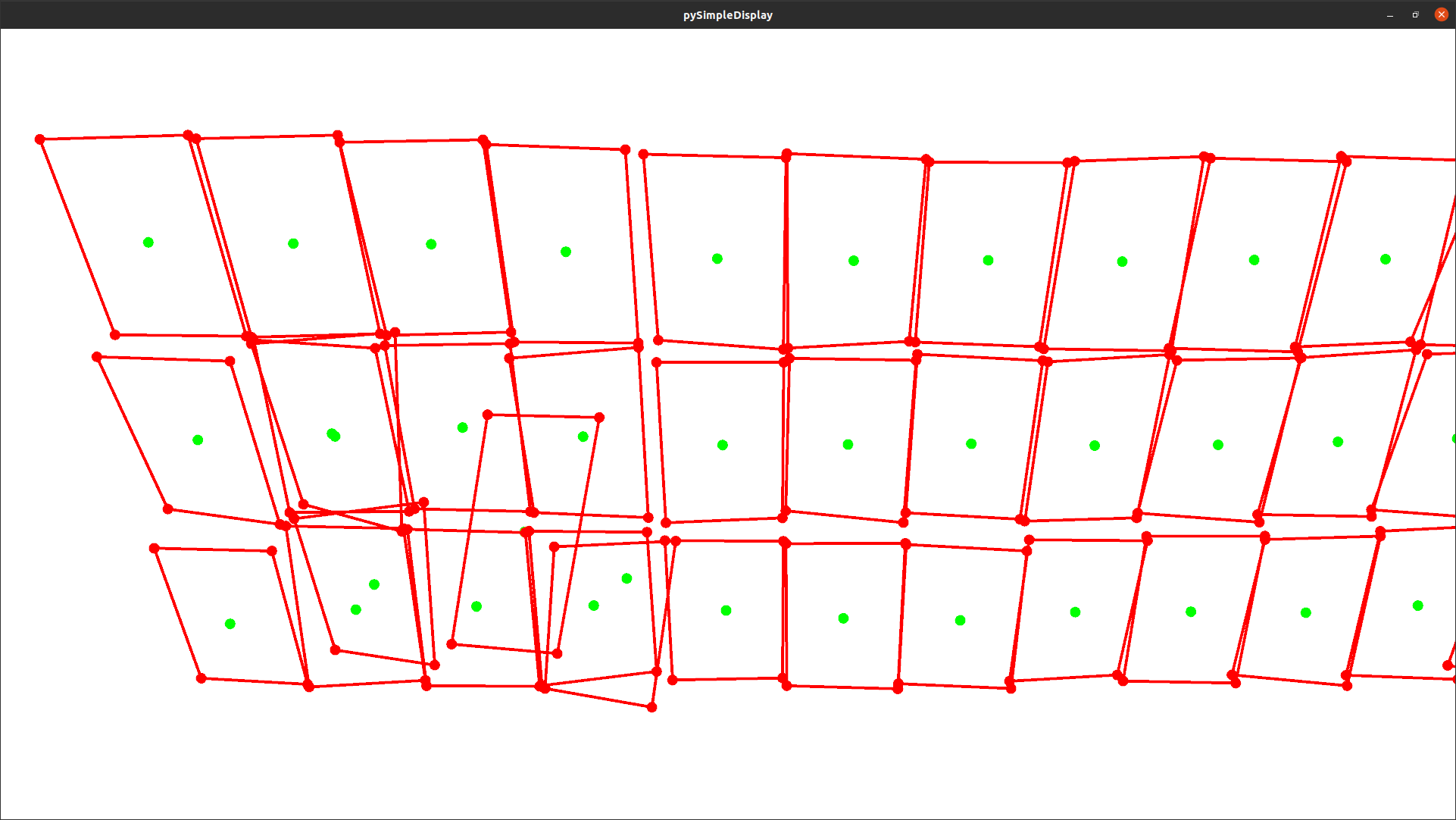}
        \caption{}
        \label{fig:module_triangulation_before_merging}
     \end{subfigure}
     \par\smallskip
     \begin{subfigure}[t]{0.49\linewidth}
         \centering
         \includegraphics[width=\linewidth, trim=1cm 3cm 0.5cm 6cm, clip]{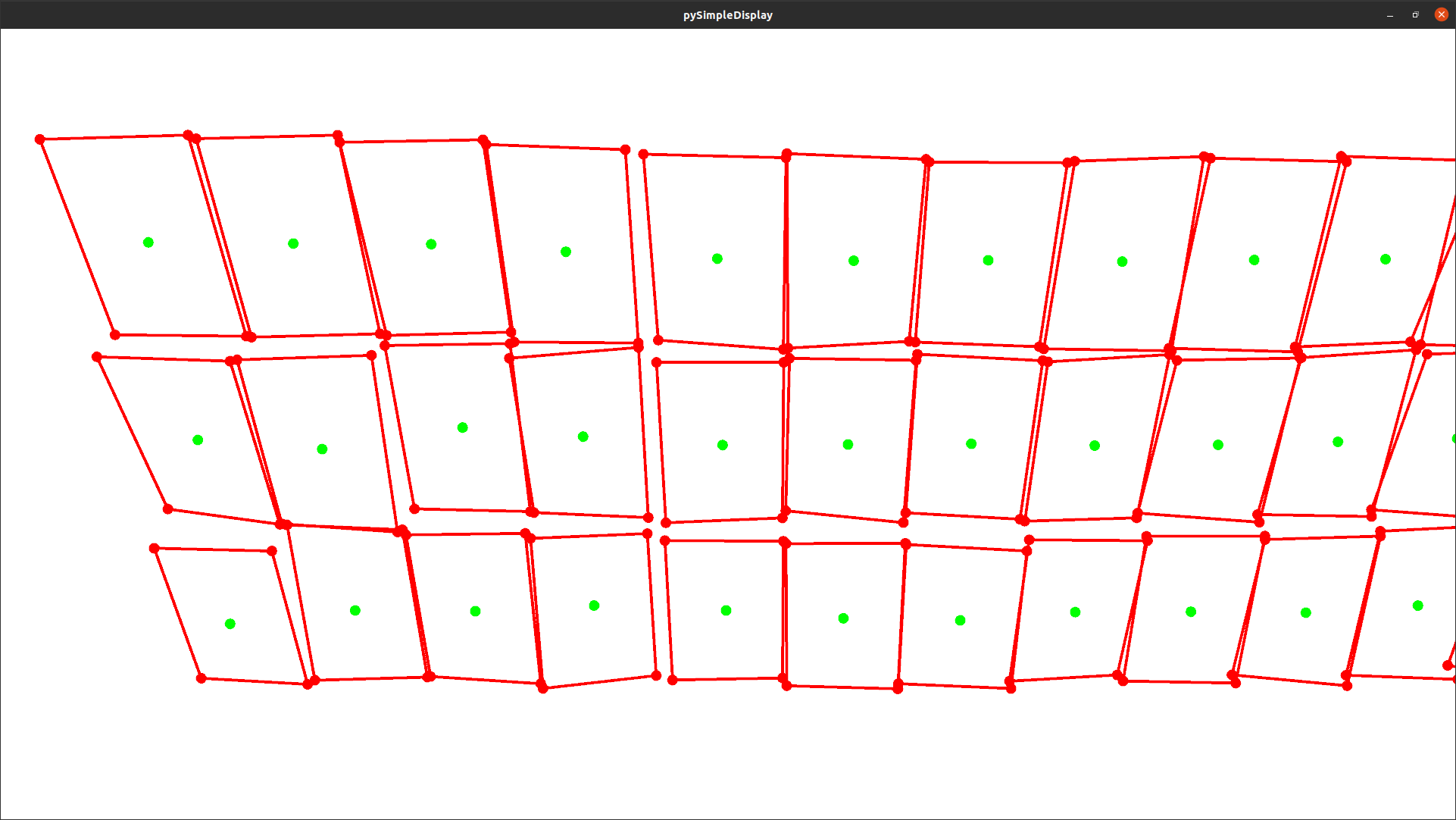}
        \caption{}
        \label{fig:module_triangulation_before_refinement}
     \end{subfigure}
     \hfill
     \begin{subfigure}[t]{0.49\linewidth}
         \centering
         \includegraphics[width=\linewidth, trim=1cm 3cm 0.5cm 6cm, clip]{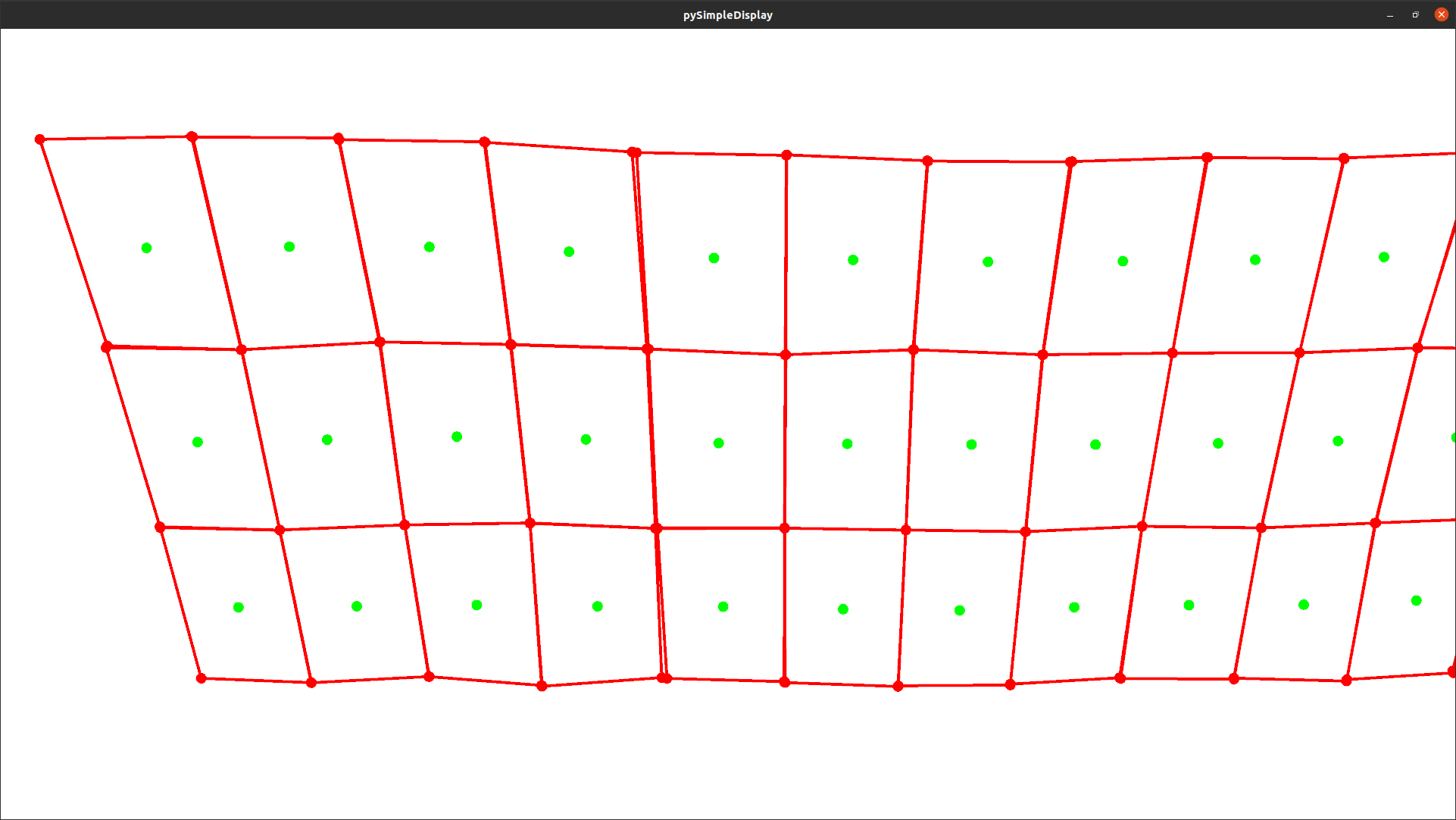}
         \caption{}
         \label{fig:module_triangulation_after_refinement}
     \end{subfigure}
        \caption{Steps of PV module triangulation: a) initial triangulation from all keyframe pairs, b) after computing median points, c) after merging duplicates, and d) after iterative refinement.}
        \label{fig:module_triangulation}
\end{figure}

The triangulation is further refined by moving nearby PV module corners closer together, yielding a smoother result (see fig.~\ref{fig:module_triangulation_after_refinement}). To this end, we build a graph containing all triangulated module points as vertices $\mathcal{P}$ and edges between those points that are close to another. Points are considered close if their Euclidean distance is at most \SI{1}{\meter} in the reconstruction and $20$ pixels in projected image coordinates. Given this graph we use the g2o graph optimization framework \cite{Kuemmerle.2011} to obtain refined module points $\mathcal{P}^*$ by optimizing the following objective
\begin{equation}
    \mathcal{P}^* = \argmin_\mathcal{P} \sum_{\left< i, j \right> \in \mathcal{C}} \rho_h \left( \mathbf{e}_{ij}^\intercal \mathbf{\Omega}_{ij} \mathbf{e}_{ij} \right) \textrm{.}
\end{equation}
Here, $\mathcal{C}$ is the set of pairs of indices for which an edge exist, $\mathbf{e}_{ij} = \mathbf{P}_i - \mathbf{P}_j$ is the difference between two points, and $\mathbf{\Omega}_{ij}$ is the information matrix, which we set to the identity matrix. The robust Huber cost function $\rho_h$ reduces the impact of outliers.

%pango.ModelViewLookAt(-53.5, 19.5, -1, -53.5, 23.5, -20, pango.AxisY), point size 15, line width 4.0

We do not apply any further refinements, such as aligning surface normals of modules, or enforcing a rectangular shape, to retain maximum flexibility of our method with respect to the layout of PV modules.

%top-down view whole plant: pango.ModelViewLookAt(50, -65, 240, 50, -65, -20, pango.AxisY)
%angled view: pango.ModelViewLookAt(-55, -35, 60, -5, 30, -40, pango.AxisZ)

%###########################################################################################

\subsection{Final Dataset Structure}

After triangulation, LTP coordinates of PV module corners and center points are transformed back to WGS\nobreakdash-$84$ coordinates and stored in a GeoJSON file together together with the module tracking ID. Similarly, the extracted IR patches of each module are stored as image files in a directory named after the tracking ID. This dataset structure allows for analysis of the extracted image patches and visualization of results on a map.

%% file: results.tex
\section{Experiments \& Results}

In this section, we apply our method to five different PV plants. We quantify the module extraction success rate together with the georeferencing error, and validate the tools' ability to process multiple plant rows in parallel. We further map predicted module anomalies and module temperatures and investigate, to what extent the temperature distribution can replace a deep learning-based classifier for the detection of abnormal modules.

%###########################################################################################

\subsection{Video Dataset}

\begin{figure}[tbp]
     \centering
     \begin{subfigure}[b]{0.32\linewidth}
         \centering
         \begin{overpic}[width=\textwidth]{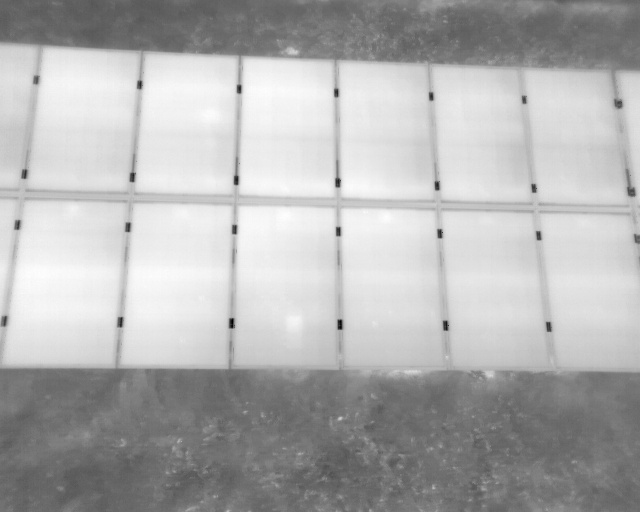}
         \put (5,5) {Plant A/B}
         \end{overpic}
     \end{subfigure}
     \begin{subfigure}[b]{0.32\linewidth}
         \centering
         \begin{overpic}[width=\textwidth]{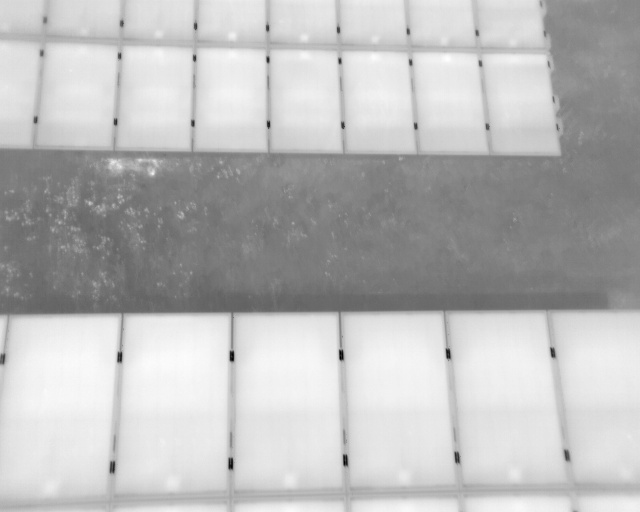}
         \put (5,5) {A/B}
         \end{overpic}
     \end{subfigure}
     \begin{subfigure}[b]{0.32\linewidth}
         \centering
         \begin{overpic}[width=\textwidth]{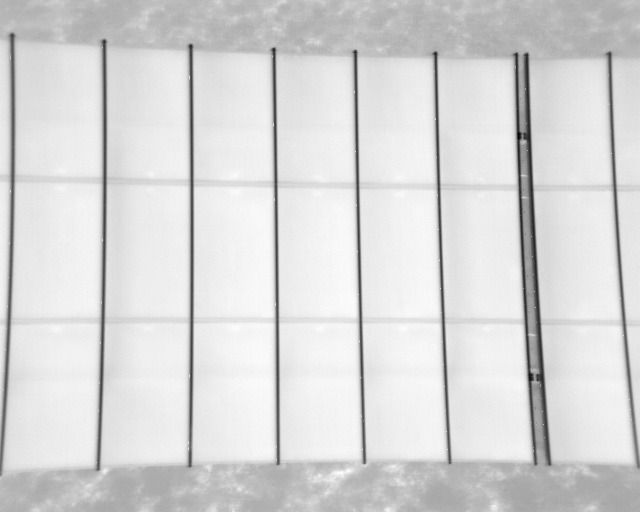}
         \put (5,5) {C/D}
         \end{overpic}
     \end{subfigure}
     \par\medskip
     \begin{subfigure}[b]{0.32\linewidth}
         \centering
         \begin{overpic}[width=\textwidth]{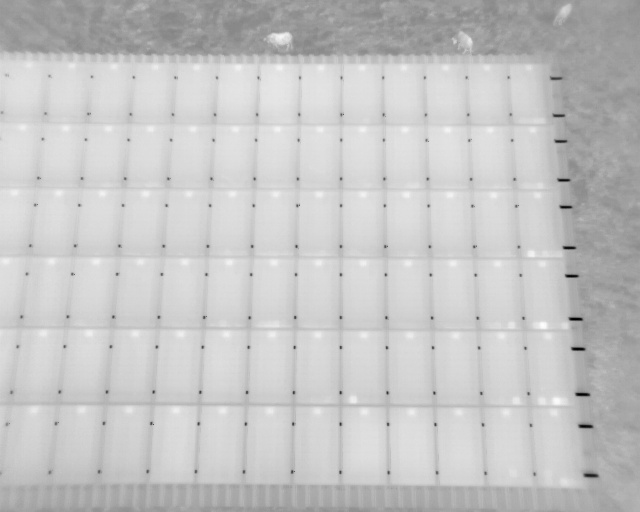}
         \put (5,5) {E}
         \end{overpic}
     \end{subfigure}
     \begin{subfigure}[b]{0.32\linewidth}
         \centering
         \begin{overpic}[width=\textwidth]{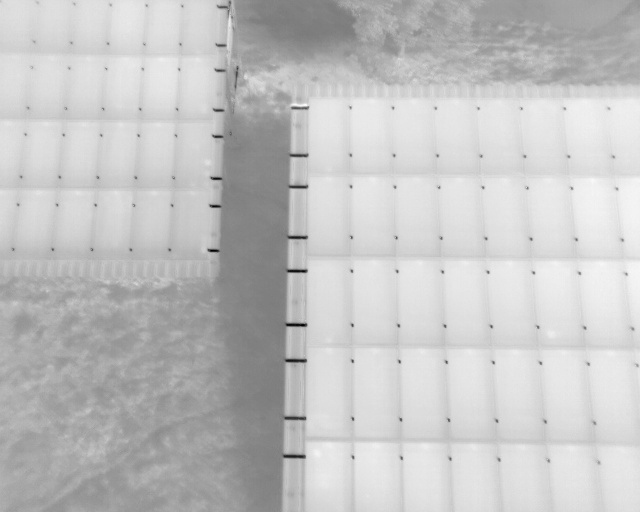}
         \put (5,5) {E}
         \end{overpic}
     \end{subfigure}
     \begin{subfigure}[b]{0.32\linewidth}
         \centering
         \begin{overpic}[width=\textwidth]{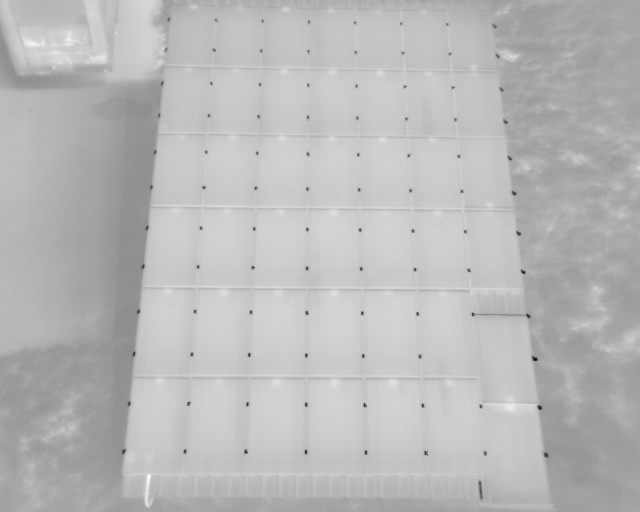}
         \put (5,5) {E}
         \end{overpic}
     \end{subfigure}
        \caption{Exemplary IR video frames of the PV plants in our study.}
        \label{fig:dataset_example_images}
\end{figure}

To validate our method, we acquire IR videos of five PV plants in Germany with a combined $35084$ PV modules using a drone of type DJI Matrice $210$. Tab.~\ref{tab:dataset_details} contains details of the PV plants, drone flights and weather conditions during data acquisition. Plants A to D are large-scale open-space plants with regular row-based layouts. Plant E is a less regular arrangement of PV arrays mounted on several rooftops. All plants consist of $60$\nobreakdash-cell crystalline silicon modules. Videos of plants A, B and E are recorded by a DJI Zenmuse XT$2$ thermal camera with $640 \times 512$ pixels resolution, \SI{8}{\hertz} frame rate and \SI{13}{\milli\meter} focal length. For plants C and D we use another variant of the DJI Zenmuse XT$2$ with \SI{30}{\hertz} frame rate and \SI{19}{\milli\meter} focal length. Fig.~\ref{fig:dataset_example_images} shows exemplary video frames from our dataset. Note, that we could not use the dataset from our previous work due to an exchange of our camera and unavailability of calibrated camera parameters.

\begin{table*}[tpb]
\centering
\caption{Details of PV plants, drone flights and weather conditions in our study. Start and end time are in UTC+2:00. Peak velocity is the \SI{99.9}{\percent} quantile of all velocities estimated from position and time delta of subsequent video frames. Weather data is from Deutscher Wetterdienst \cite{DWD.2021}. We report mean and standard deviation of measurements taken at the nearest weather station every $10$ minutes during the flight.}
\label{tab:dataset_details}
\begin{tabular}{l
S[table-format=5.0]
l
l
l
S[table-format=5.0]
S[table-format=4.0]@{\,}l
S[table-format=1.1]@{\,}l
S[table-format=2.1,table-figures-uncertainty=1,separate-uncertainty=true]@{\,}l
S[table-format=2.1,table-figures-uncertainty=3,separate-uncertainty=true]@{\,}l
S[table-format=1.1,table-figures-uncertainty=1,separate-uncertainty=true]@{\,}l
l}
\toprule
\multicolumn{3}{c}{Plant Details} & \multicolumn{7}{c}{Flight Details} & \multicolumn{7}{c}{Weather Conditions}\\\cmidrule(lr){1-3}\cmidrule(lr){4-10}\cmidrule(lr){11-17}
{ID} & {\# Modules} & {Type} & {Start time} & {End time} & {\# Frames} & \multicolumn{2}{c}{Distance} & \multicolumn{2}{c}{Peak velocity} & \multicolumn{2}{c}{Air temp.} & \multicolumn{2}{c}{Global radiation} & \multicolumn{2}{c}{Wind speed} & {Wind dir.}\\\midrule
A & 13640 & open-space & 10:28:48 & 12:40:14 & 42272 & 7612 & \si{\meter} & 4.1 & \si{\meter\per\second} & 25.9(5) & \si{\celsius} & 39.7(18) & \si{\joule\per\centi\meter\squared} & 2.8(4) & \si{\meter\per\second} & WSW\\
B & 5280 & open-space & 13:37:59 & 14:14:30 & 13715 & 2929 & \si{\meter} & 4.1 & \si{\meter\per\second} & 26.8(4) & \si{\celsius} & 30.3(61) & \si{\joule\per\centi\meter\squared} & 3.6(4) & \si{\meter\per\second} & SW\\
C & 6210 & open-space & 12:16:21 & 12:39:05 & 34593 & 2468 & \si{\meter} & 6.6 & \si{\meter\per\second} & 22.3(3) & \si{\celsius} & 46.2(106) & \si{\joule\per\centi\meter\squared} & 5.7(6) & \si{\meter\per\second} & WNW\\
D & 8460 & open-space & 11:01:00 & 11:33:34 & 50348 & 3479 & \si{\meter} & 7.2 & \si{\meter\per\second} & 23.4(2) & \si{\celsius} & 57.4(15) & \si{\joule\per\centi\meter\squared} & 2.0(8) & \si{\meter\per\second} & W\\
E & 1494 & rooftops & 11:30:24 & 11:54:02 & 4527 & 485 & \si{\meter} & 4.2 & \si{\meter\per\second} & 19.0(3) & \si{\celsius} & 42.8(30) & \si{\joule\per\centi\meter\squared} & 2.6(3) & \si{\meter\per\second} & SE\\
\bottomrule
\end{tabular}
\end{table*}

%###########################################################################################

\subsection{Module Extraction Success Rate}

We apply our method on the five PV plants in our video dataset and extract over $2.2$ million IR images of the $35084$ PV modules of all plants (on average $64.1$ images per module). Additionally, geocoordinates are obtained for each module as exemplary shown for plant B in fig.~\ref{fig:module_layout_plant_B} (for the other plants see appendix~\ref{sec:additional_georeferencing_results}). As detailed in tab.~\ref{tab:quantitative_extraction_results}, \SI{99.3}{\percent} of all modules are successfully extracted and georeferenced. As compared to the \SI{87.8}{\percent} success rate of our previous work, we now miss only one in $140$ modules instead of one in eight. This $18$\nobreakdash-fold improvement of the extraction success rate is mostly due to the higher robustness of our new method to errors in the data acquisition process, such as cropping of the scanned row, double acquisition of the same row, or small loops in the drone trajectory. Such an error caused our previous method to loose all modules in an entire plant row. Opposed to that, our new method can handle many of those acquisition errors, and fails at most locally for a few modules. The almost perfect success rate of our new method is important in practice, as every missed module is a missed opportunity to increase yield and profitability of the plant. Furthermore, safety critical anomalies (e.g. fire hazards) could be overlooked.

Tab.~\ref{tab:quantitative_extraction_results} also contains a detailed breakdown of the failure modes of the $234$ modules missed by our method. In total, $13$ modules exhibit substantial distortions, and $40$ modules are missing in the reconstructions, because they are not covered by sufficiently many video frames to be accurately triangulated. Another $181$ modules appear multiple times in the reconstruction because the merging procedure (sec.~\ref{sec:merging_of_duplicate_detections}) failed. This happens for modules appearing in video frames, which are temporally far apart. As these frames have a large relative pose error (due to the use of standard GPS) the triangulated modules do not align well and can not be merged correctly. This can most likely be mitigated by using RTK-GPS. Finally, there are $24$ false positive modules corresponding to other objects, which are mistaken as PV modules by the Mask R-CNN segmentation model.

\begin{figure}[tbp]
    \centering
    \includegraphics[height=\linewidth, angle=90]{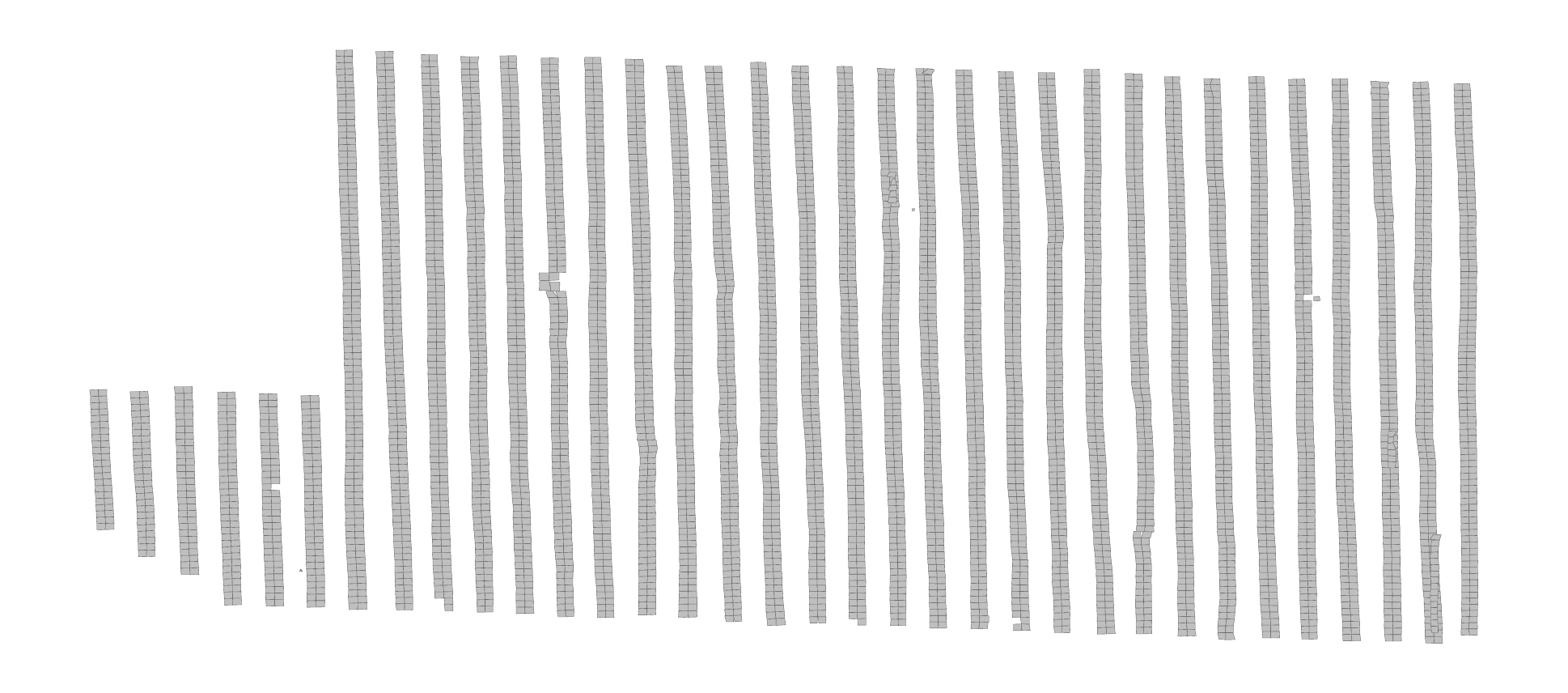}
    \caption{Map with estimated geocoordinates of PV modules in plant B.}
    \label{fig:module_layout_plant_B}
\end{figure}

\begin{table}[tpb]
\centering
\caption{Numbers of PV modules and module image patches extracted from the plants in our dataset. Failures are missing (MM), duplicate (DP), distorted (DS) and false positive (FP) modules.}
\label{tab:quantitative_extraction_results}
\begin{tabular}{l
% patches
S[table-format=5.0]
S[table-format=5.0]
S[table-format=7.0]
S[table-format=3.1]
S[table-format=2.0]
S[table-format=3.0]
S[table-format=1.0]
S[table-format=2.0]}
\toprule
{Plant} & \multicolumn{2}{c}{\# Modules} & \multicolumn{2}{c}{\# Patches} & \multicolumn{4}{c}{\# Failures}\\\cmidrule(lr){2-3}\cmidrule(lr){4-5}\cmidrule(lr){6-9}
 & {Total} & {Extracted} & {Extracted} & {$\varnothing$/Module} & {MM} & {DP} & {DS} & {FP}\\\midrule
A & 13640 & 13463 & 398221 & 29.2 & 18 & 152 & 7 & 13\\
B & 5280 & 5246 & 140120 & 26.6 & 6 & 28 & 0 & 3\\
C & 6210 & 6200 & 635437 & 102.3 & 4 & 1 & 5 & 2\\
D & 8460 & 8453 & 936867 & 110.8 & 7 & 0 & 0 & 4\\
E & 1494 & 1488 & 138008 & 90.8 & 5 & 0 & 1 & 2\\\midrule
Total & 35084 & 34850 & 2248653 & 64.1 & 40 & 181 & 13 & 24\\
\bottomrule
\end{tabular}
\end{table}
% number of patches: run "find . -type f | wc -l" in patches/radiometric

%###########################################################################################

\subsection{Georeferencing Accuracy}

In this section we quantify the accuracy of the georeferenced PV module locations in terms of the root mean square error (RMSE) between estimated LTP geocoordinates $(\hat{e}, \hat{n})$ and ground truth geocoordinates $(e, n)$ of $N$ selected PV modules
\begin{equation}
    \textrm{RMSE} = \sqrt{ \frac{1}{N} \sum_{i=1}^{N} \left( \hat{e}_i - e_i \right)^2 + \frac{1}{N} \sum_{i=1}^{N} \left( \hat{n}_i - n_i \right)^2 } \textrm{.}
\end{equation}
Here, $e$ and $n$ are the east and north positions in the LTP coordinate system. The altitude coordinate is omitted as for mapping only the horizontal error is of interest. Because point correspondences have to be found manually, we select only every $11$th module in every second row of the plant and consider only the top-left module corners. We further limit the accuracy analysis to plant A, as it is the largest plant in our dataset and the only one, for which a ground truth is available. Ground truth positions are obtained from an orthophoto of the plant. This is possible, as this orthophoto exhibits a small RMSE of less than \SI{2}{\centi\meter}, facilitated by the use of RTK-GPS, ground control points, high-resolution visual imagery and a higher flight altitude.

\begin{figure}[tbp]
    \centering
    \includegraphics[height=\linewidth, angle=90]{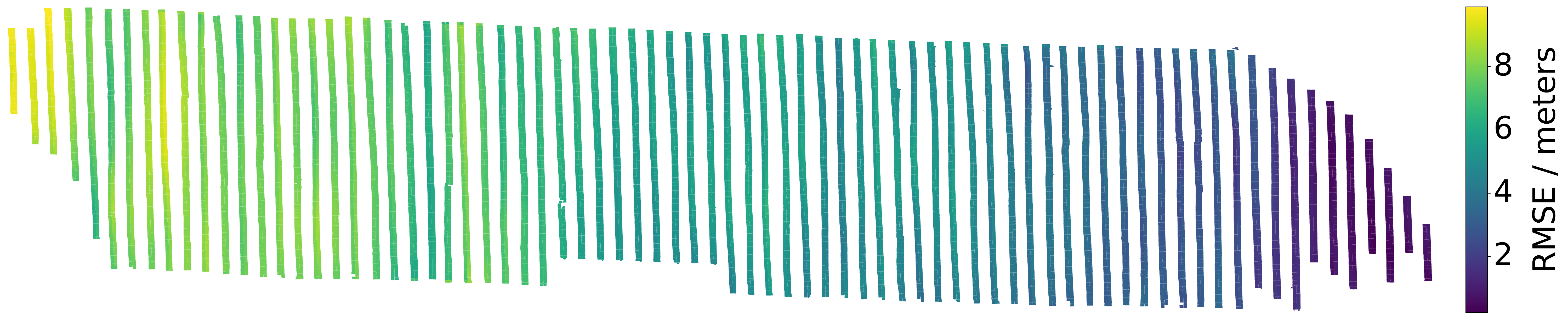}
    \caption{RMSE between ground truth and estimated horizontal geocoordinates of PV modules in plant A. The RMSE is computed for selected points (black dots) and linearly interpolated on a $4000 \times 800$ grid (heatmap).}
    \label{fig:rmse_map}
\end{figure}

The RMSE for the entire plant is \SI{5.87}{\meter}. This is close to the expected \SI{4.9}{\meter} accuracy of GPS under open sky conditions \cite{Diggelen.2015}. However, the RMSE is not constant for the entire plant, but instead smoothly increases from \SI{0.42}{\meter} in the east to \SI{9.39}{\meter} in the west. This also becomes evident in fig.~\ref{fig:rmse_map}, which shows the spatial interpolation of the RMSE over the entire plant. This error drift in the SfM reconstruction is most likely caused by the low accuracy and unknown DOP of the measured GPS trajectory of the drone. As the SfM reconstruction consist of seven partial reconstructions and we do not use ground control points, another possible cause is misalignment of the partial reconstructions.

To analyze the distortion of each individual row, we remove the trend in the RMSE distribution. To this end, we align each row with the respective ground truth positions prior to computing the RMSE for the row. The resulting per-row RSME values range from \SI{0.22}{\meter} to \SI{0.82}{\meter}, indicating low distortion of individual rows. Due to this, accurate localization of PV modules within the plant is possible, despite the large absolute RMSE of \SI{5.87}{\meter}.

%###########################################################################################

\subsection{Simultaneous Processing of Multiple Rows}

\begin{figure}[tbp]
     \centering
     \begin{subfigure}[b]{0.32\linewidth}
         \centering
         \includegraphics[width=\textwidth]{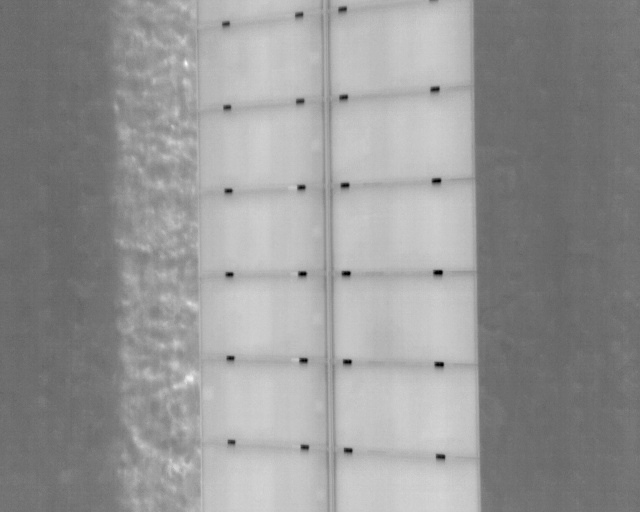}
     \end{subfigure}
     \begin{subfigure}[b]{0.32\linewidth}
         \centering
         \includegraphics[width=\textwidth]{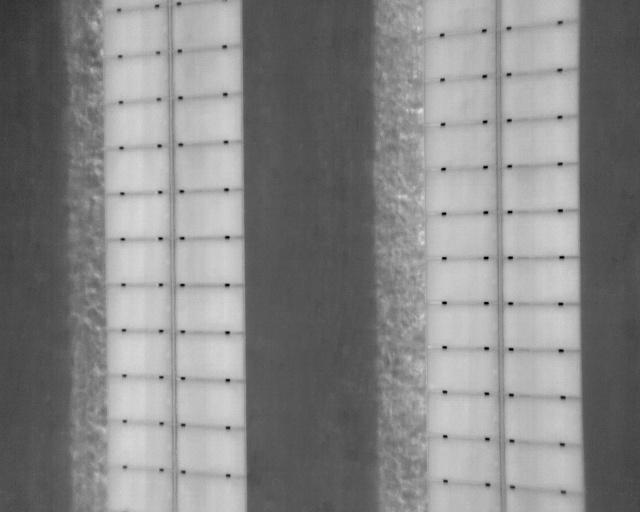}
     \end{subfigure}
     \begin{subfigure}[b]{0.32\linewidth}
         \centering
         \includegraphics[width=\textwidth]{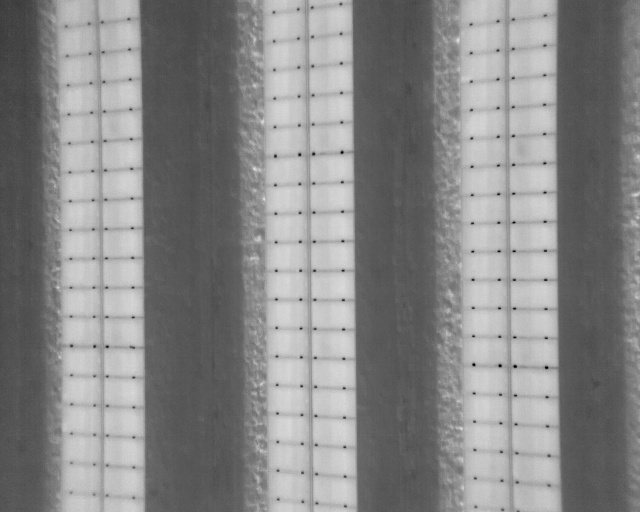}
     \end{subfigure}
    \par\medskip
     \begin{subfigure}[b]{0.32\linewidth}
         \centering
         \includegraphics[width=\textwidth, trim=3cm 0.3cm 4.3cm 1.7cm, clip]{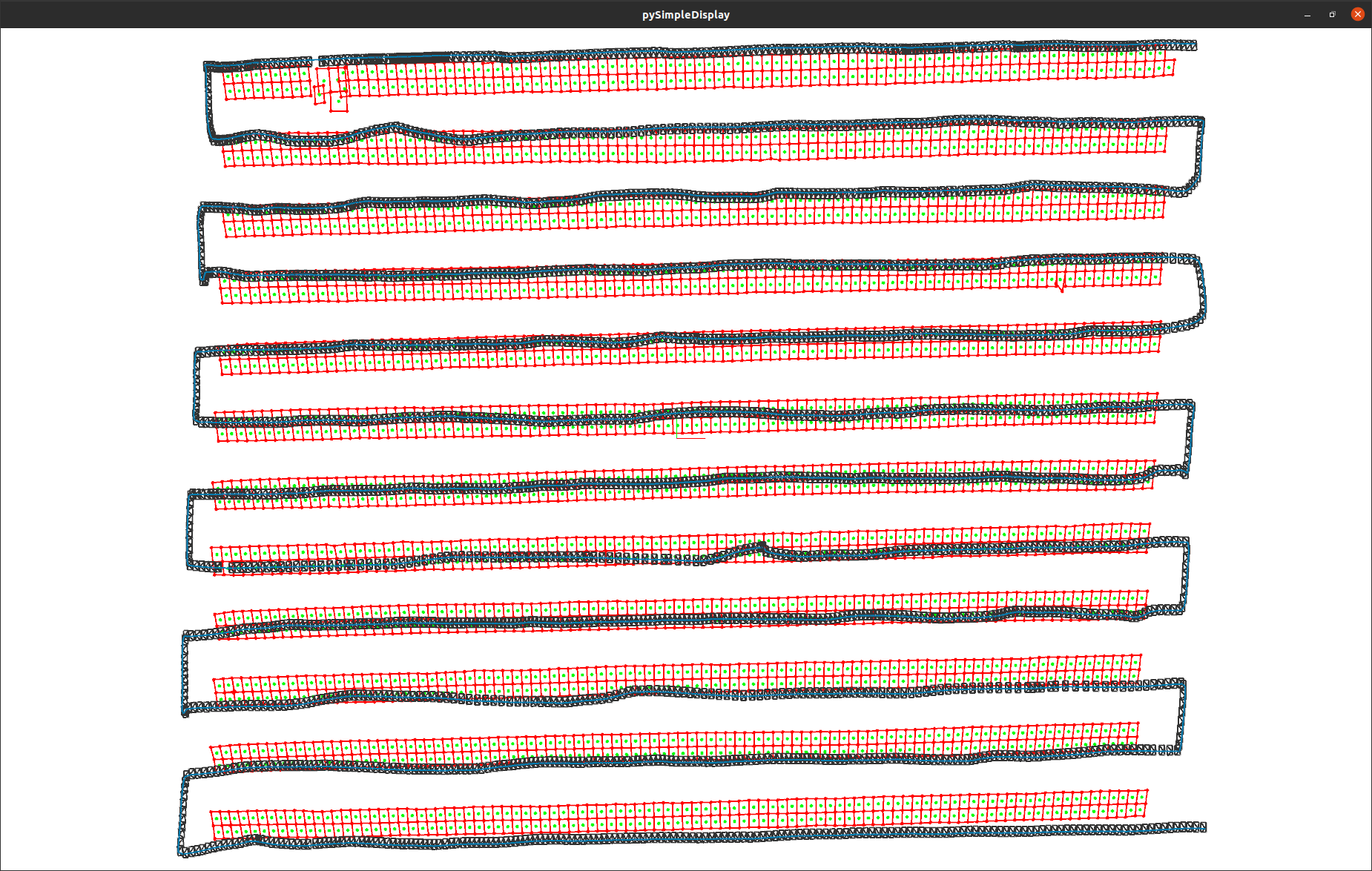}
     \end{subfigure}
     \begin{subfigure}[b]{0.32\linewidth}
         \centering
         \includegraphics[width=\textwidth, trim=3cm 0.3cm 4.3cm 1.7cm, clip]{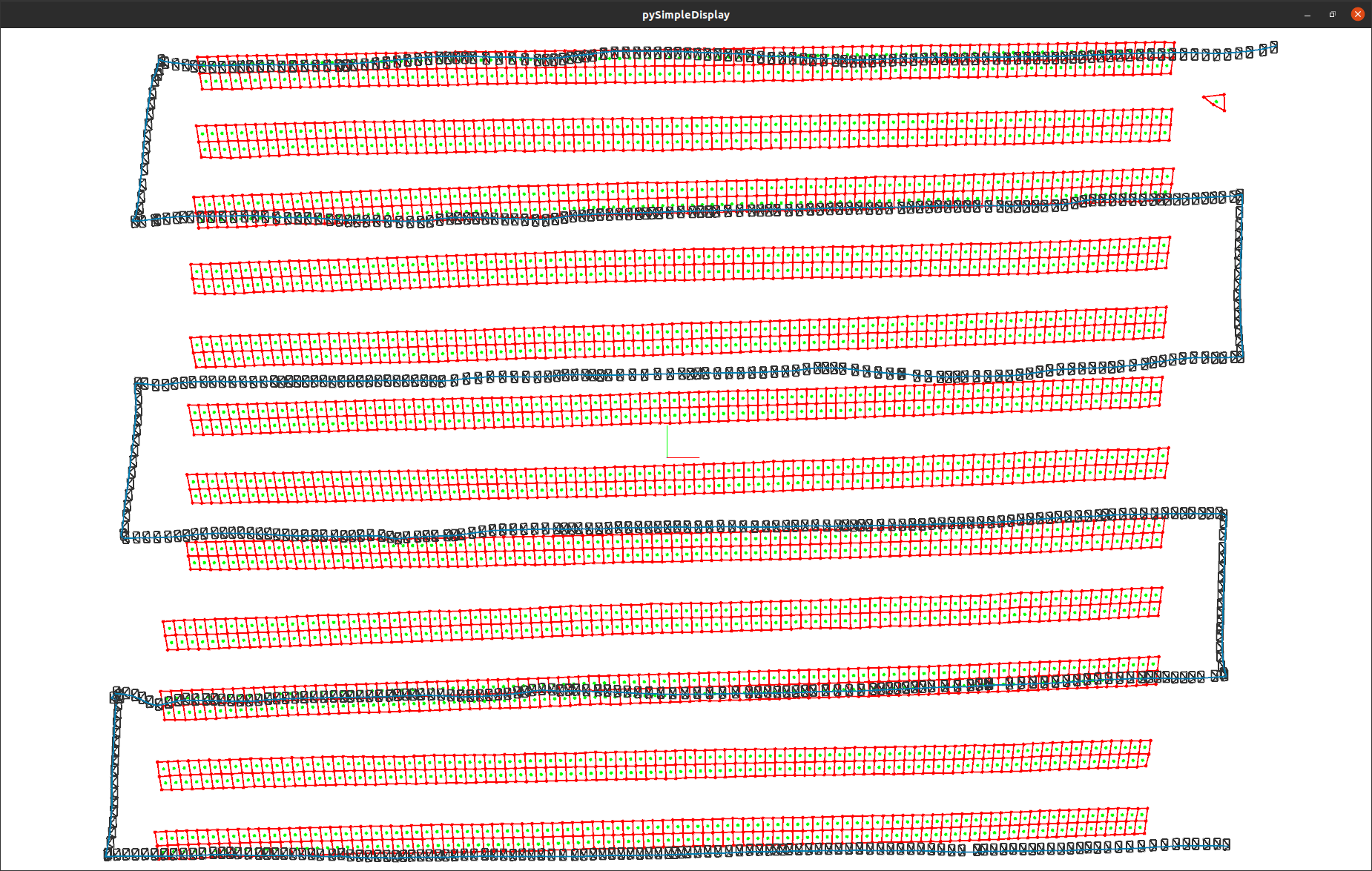}
     \end{subfigure}
     \begin{subfigure}[b]{0.32\linewidth}
         \centering
         \includegraphics[width=\textwidth, trim=3cm 0.3cm 4.3cm 1.7cm, clip]{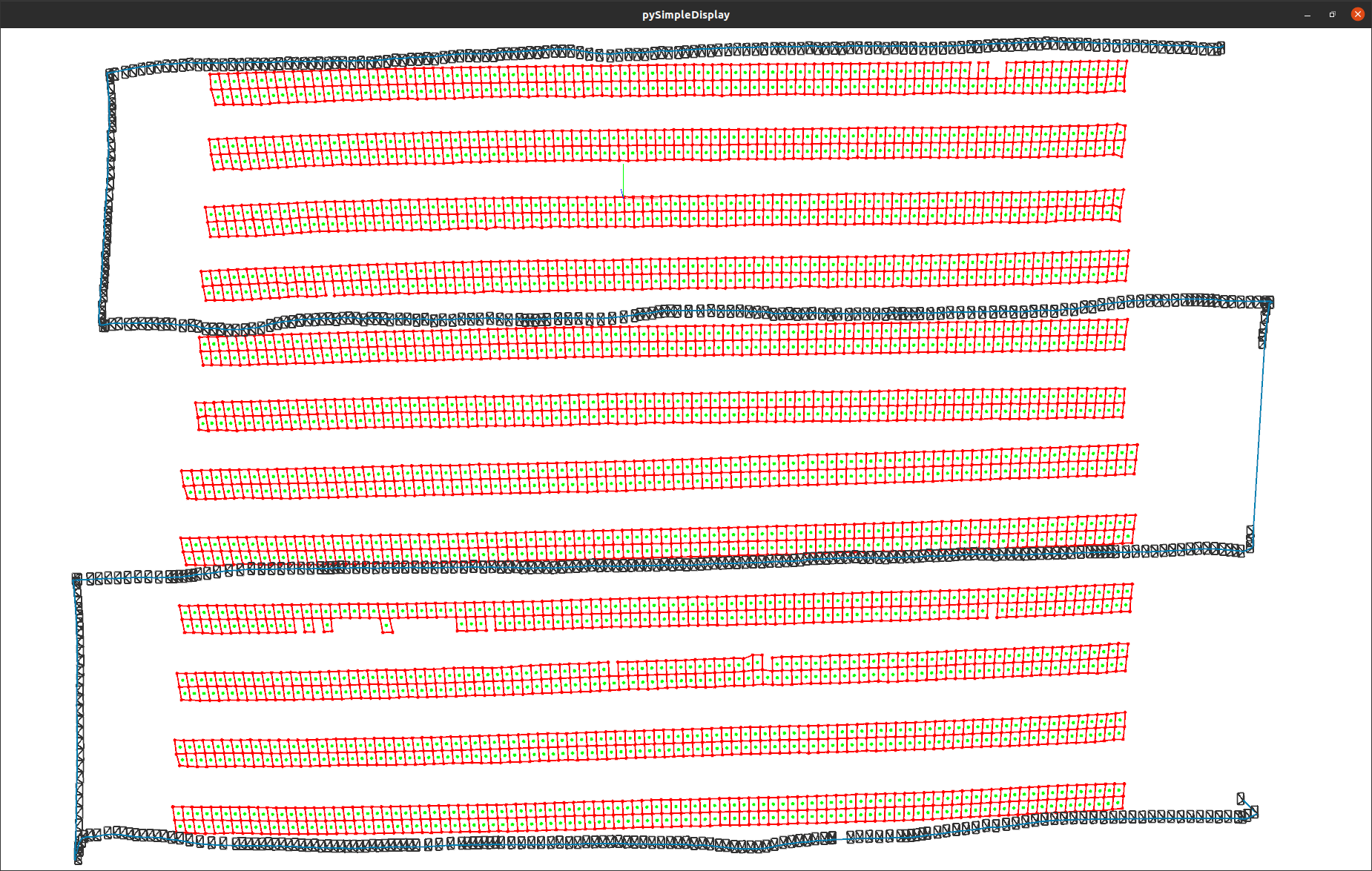}
     \end{subfigure}
        \caption{Top row: Exemplary video frames for scanning one, two and three PV plant rows simultaneously. Bottom row: Resulting reconstructions of modules and flight trajectory.}
        \label{fig:simultaneous_processing_of_rows}
\end{figure}

One important advantage of our new method is the ability to process multiple PV plant rows simultaneously. We validate this experimentally by acquiring IR videos of the first $12$ rows of plant A. We perform three flights, scanning one, two and three rows at a time. Fig.~\ref{fig:simultaneous_processing_of_rows} shows exemplary video frames of each flight as well as the reconstructions of modules and flight trajectories produced by our method.

As reported in tab.~\ref{tab:results_simultaneous_processing_of_rows}, scanning two and three rows simultaneously speeds up the flight duration by a factor of $2.1$ and $3.7$, respectively. Module throughput increases accordingly from \SI{3.36}{\per\second} to \SI{7.03}{\per\second} and \SI{12.57}{\per\second}. This means, scanning all $2376$ modules of the $12$ selected rows takes only $338$ or $189$ seconds when scanning two or three rows at a time. Additionally, flight distance decreases by a factor of $1.9$ and $2.8$. This has the benefit of increasing the range of the drone before a battery change is needed. The cost for the improvement in throughput is a two- or threefold reduction in the resolution of extracted module images. Furthermore, we found the manual flight is slightly more complicated when scanning three rows at a time instead of one or two, because it is easier to miscount the rows when shifting over to the next row triplet. However, this is not a limitation when flying autonomously.

This experiment confirms the ability of our method to significantly increase throughput simply by scanning more than one plant row at a time. This is highly relevant in practice, as it significantly reduces duration and cost of the inspection. It is also an improvement over our previous method \cite{Bommes.2021}, which could process only one row at a time.

\begin{table}[tpb]
\centering
\caption{Results for simultaneous scanning of one, two and three rows.}
\label{tab:results_simultaneous_processing_of_rows}
\begin{tabular}{llll}
\toprule
{} & {One Row} & {Two Rows} & {Three Rows}\\\midrule
Flight distance & \SI{1307}{\meter} & \SI{681}{\meter} & \SI{461}{\meter}\\
Flight duration & \SI{707}{\second} & \SI{338}{\second} & \SI{189}{\second}\\
Average module resolution & \SI{141}{\pixel} \texttimes{} \SI{99}{\pixel} & \SI{73}{\pixel} \texttimes{} \SI{50}{\pixel} & \SI{46}{\pixel} \texttimes{} \SI{33}{\pixel}\\
Module throughput & \SI{3.36}{\per\second} & \SI{7.03}{\per\second} & \SI{12.57}{\per\second}\\
\bottomrule
\end{tabular}
\end{table}
% Missed modules & 1 & 0 & 22 \\
% additional metric: Defect detection AUROC, processing time

%###########################################################################################

\subsection{Mapping Module Anomalies}
\label{sec:mapping_module_defects}

\begin{figure}[tbp]
    \centering
    \includegraphics[width=\linewidth]{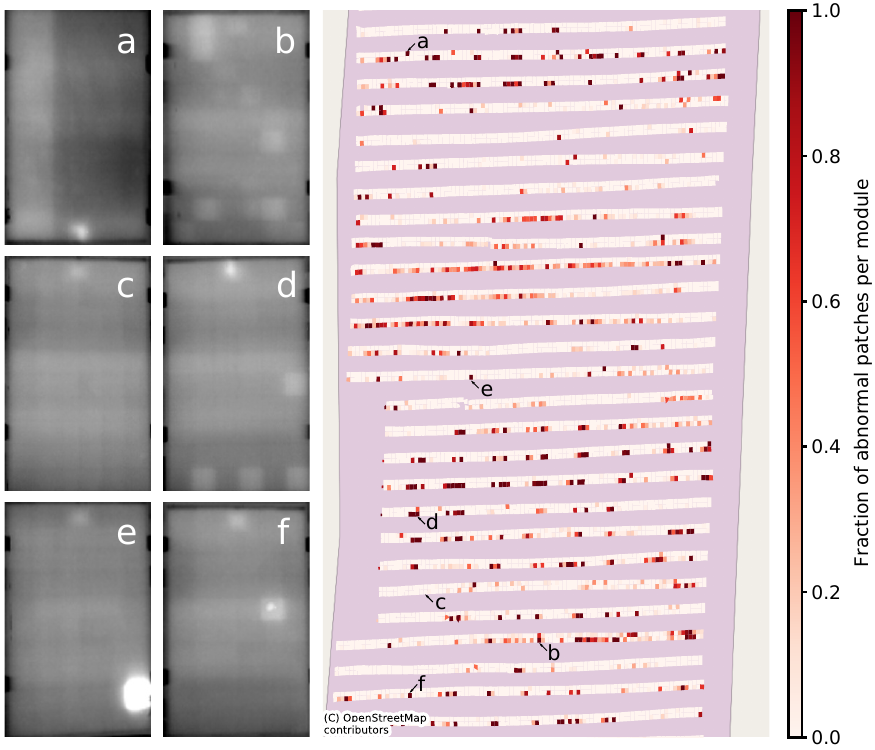}
    \caption{Map of predicted module anomalies in a section of plant A. On the left, IR images of the modules highlighted on the map are shown. Their temperature range is \SI{30}{\celsius} (black) to \SI{50}{\celsius} (white).}
    \label{fig:module_defects_map}
\end{figure}

In this section we apply a deep learning-based binary classifier to the extracted IR image patches of each PV module in plant A, which predicts whether the module is abnormal or not. \footnote{We use the ResNet\nobreakdash-$34$ convolutional neural network (CNN) classifier from Bommes et al.~\cite{Bommes.2021b}, which is trained with a supervised cross-entropy loss on labelled IR module patches of plant B in the dataset of the original work.} We then use the estimated module geocoordinates to visualize the distribution of abnormal modules on a map (see fig.~\ref{fig:module_defects_map}). Since there are multiple images for each module, we can plot the fraction of images, in which a module is predicted as abnormal. We call this the \emph{anomaly ratio}. As opposed to a simple binary prediction, the anomaly ratio is an approximate indicator for the severity of a module anomaly. This is, because for severe, i.e. clearly visually expressed, anomalies the classifier is more confident, reaching a larger consensus of its predictions over all images of a module.

The so obtained anomaly map enables not only targeted repairs of severely abnormal modules, but also facilitates the identification of fundamental problems of the plant. For the analyzed plant, we find for example, that anomalies occur much more frequently in the bottom row, where modules are closer to the ground, rather than in the top row. A possible explanation for this is the intrusion of moisture into the PV modules near the ground. Being aware of such an issue allows the operator to monitor affected modules more thoroughly and to take action to prevent further damage to the plant.

%###########################################################################################

\subsection{Mapping Module Temperatures}
\label{sec:mapping_module_temperatures}

\begin{figure}[tbp]
     \captionsetup[subfigure]{aboveskip=2pt, belowskip=0pt, justification=centering}
     \centering
     \begin{subfigure}[t]{0.32\linewidth}
         \centering
         \includegraphics[scale=0.21]{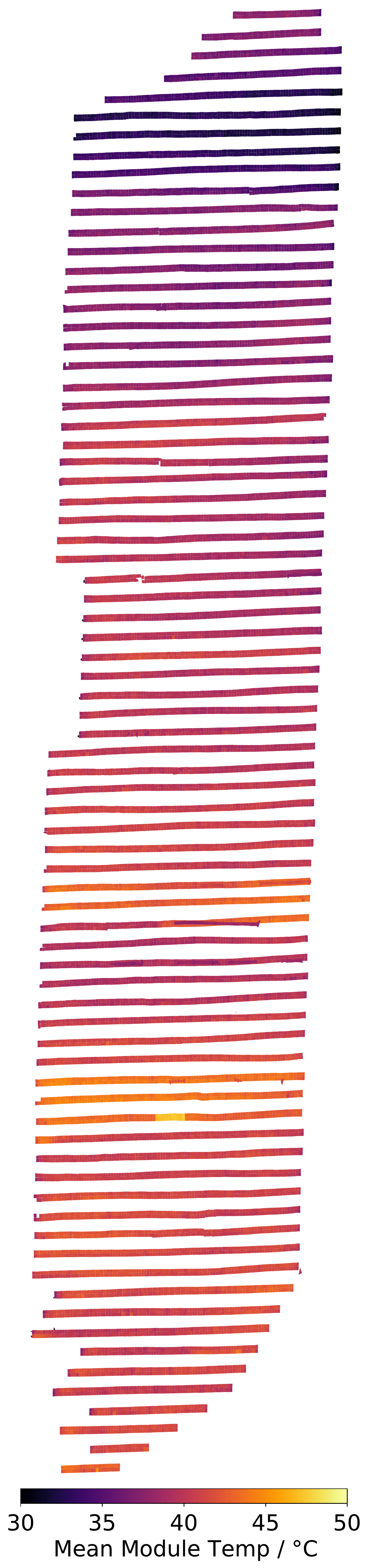}
        \caption{}
        \label{fig:mean_of_mean_temps}
     \end{subfigure}
     \hfill
     \begin{subfigure}[t]{0.32\linewidth}
         \centering
         \includegraphics[scale=0.21]{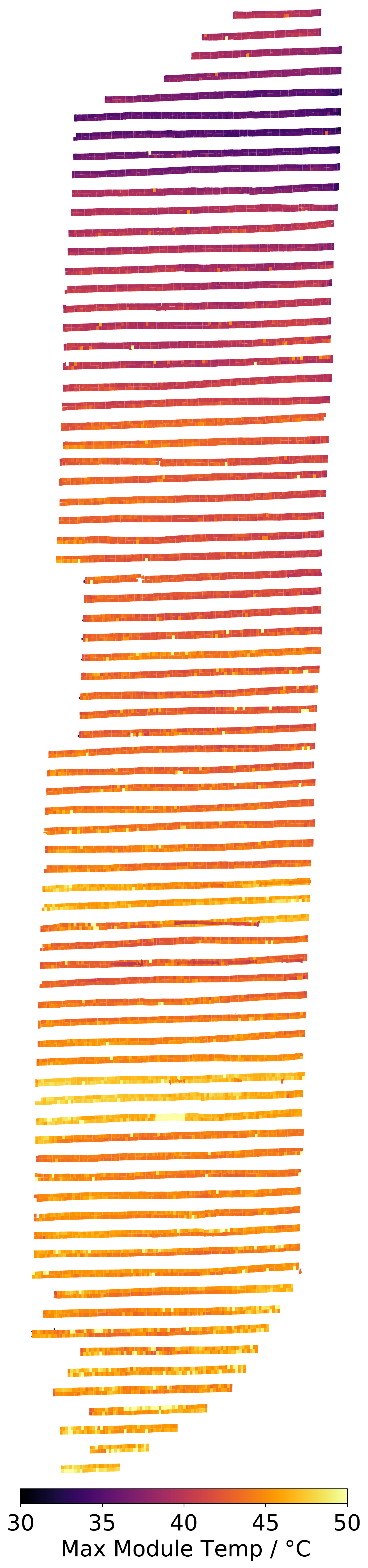}
        \caption{}
        \label{fig:mean_of_max_temps}
     \end{subfigure}
     \hfill
     \begin{subfigure}[t]{0.32\linewidth}
         \centering
         \includegraphics[scale=0.21]{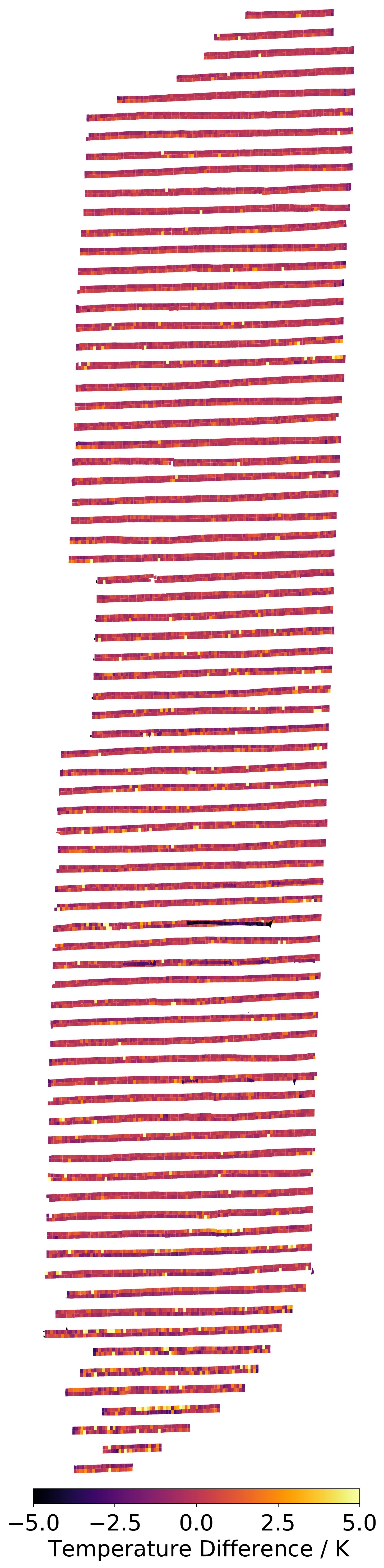}
        \caption{}
        \label{fig:mean_of_max_temps_corrected}
     \end{subfigure}
        \caption{Map of plant A showing the distribution of mean (a) and maximum (b) module temperatures. Each value is the average over all images showing a module. For higher contrast temperatures are clipped below \SI{30}{\celsius} and above \SI{50}{\celsius}. In (c) local differences of the maximum module temperature are emphasized by subtracting the median temperatures of neighbouring modules within a radius of \SI{7}{\meter}.}
        \label{fig:temperature_distributions}
\end{figure}

Apart from module anomalies, we visualize the spatial distribution of module temperatures in plant A (see fig.~\ref{fig:temperature_distributions}). Temperatures are obtained from the extracted IR image patches of each module and plotted on a map using the module geocoordinates. For each module the maximum, minimum, mean or median temperature over the module area can be computed. Prior to this, we cut away a few pixels (\SI{5}{\percent} of the image width) from the image borders to ignore module frames and mounting brackets. To obtain a final temperature value for each module, we take the mean over the values estimated for each of the image patches of the module. As opposed to using a single representative image patch or the maximum over all patches, the mean is more robust to artifacts, which may be present in some of the module images. Of both mean and maximum temperature distributions, we find the maximum temperatures (see fig.~\ref{fig:mean_of_max_temps}) more informative as they are sensitive to the local hot spots typically occurring in abnormal modules. However, both mean and maximum module temperatures reveal the global temperature distribution of the plant, which is not constant, but exhibits a low-frequency pattern with temperature differences of up to \SI{15}{\kelvin}. As module images are acquired over a duration of $132$ minutes, possible explanations for this pattern are slow changes in the solar irradiance \cite{Rajendran.2016}, cloud cover \cite{Nomura.2017, Lappalainen.2016}, air temperature, wind speed, and camera temperature \cite{Wan.2021}. The temperature distribution is also affected by local differences in the radiative and convective heat transfer, and by the number of neighbouring modules, leading to cooler modules at the edges of each plant row \cite{Denz.2020}. Direct use of this temperature distribution for anomaly detection is not possible, as there is no common threshold value, which separates normal from abnormal modules. To account for this, we compute local temperature differences between neighbouring modules. Specifically, we subtract the median of the maximum module temperatures of all neighbouring modules within a radius of \SI{7}{\meter} from each module. Fig.~\ref{fig:mean_of_max_temps_corrected} shows the resulting relative maximum module temperatures. These relative temperatures are independent of the changes in environmental conditions during the flight, and consequently facilitate detection of abnormal modules by selecting a suitable temperature threshold.

Apart from locally overheated modules, the temperature distribution allows to identify string anomalies. For example in plant A, there is an inactive string (in the middle of the $19$th row counted from the bottom), which is clearly visible in the temperature map (see fig.~\ref{fig:temperature_distributions}). Being able to identify such anomalies is important, as an entire inactive string causes large yield and power losses.

%###########################################################################################

\subsection{Anomaly Detection with Module Temperatures}

\begin{figure}[tbp]
     \captionsetup[subfigure]{font=scriptsize, aboveskip=2pt, belowskip=0pt, labelformat=empty, justification=centering}
     \newcommand\sizefactor{0.193}
     \centering
     \begin{subfigure}[t]{\sizefactor\columnwidth}
         \centering
         \includegraphics[width=\textwidth]{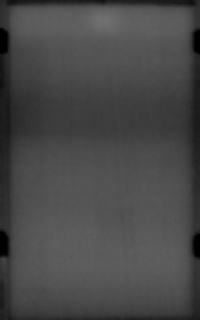}
        \caption{Healthy}
     \end{subfigure}
     \begin{subfigure}[t]{\sizefactor\columnwidth}
         \centering
         \includegraphics[width=\textwidth]{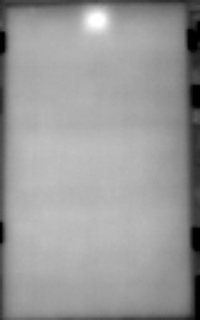}
         \caption{Mh: Module open-circuit}
     \end{subfigure}
     \begin{subfigure}[t]{\sizefactor\columnwidth}
         \centering
         \includegraphics[width=\textwidth]{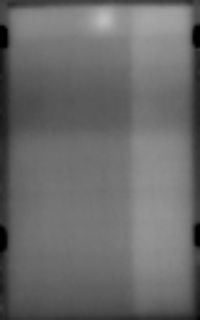}
         \caption{Sh: Substring open-circuit}
     \end{subfigure}
     \begin{subfigure}[t]{\sizefactor\columnwidth}
         \centering
         \includegraphics[width=\textwidth]{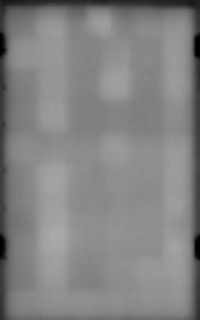}
         \caption{Pid: Potential- induced degrad.}
     \end{subfigure}
     \begin{subfigure}[t]{\sizefactor\columnwidth}
         \centering
         \includegraphics[width=\textwidth]{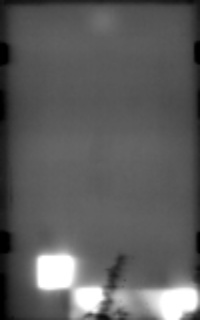}
         \caption{Cm+: Multiple hot cells}
     \end{subfigure}
     \par\smallskip
     \begin{subfigure}[t]{\sizefactor\columnwidth}
         \centering
         \includegraphics[width=\textwidth]{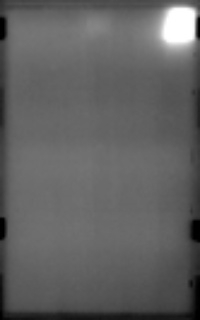}
         \caption{Cs+: Single hot cell}
     \end{subfigure}
     \begin{subfigure}[t]{\sizefactor\columnwidth}
         \centering
         \includegraphics[width=\textwidth]{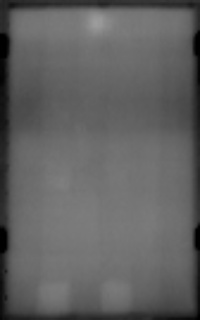}
         \caption{Cs/Cm: Warm cell(s)}
     \end{subfigure}
     \begin{subfigure}[t]{\sizefactor\columnwidth}
         \centering
         \includegraphics[width=\textwidth]{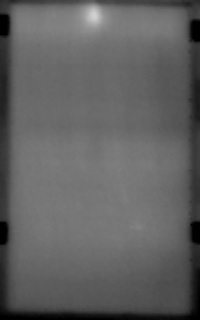}
         \caption{D: Diode overheated}
     \end{subfigure}
     \begin{subfigure}[t]{\sizefactor\columnwidth}
         \centering
         \includegraphics[width=\textwidth]{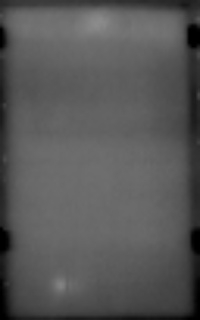}
         \caption{Chs: Hot spot}
     \end{subfigure}
     \begin{subfigure}[t]{\sizefactor\columnwidth}
         \centering
         \includegraphics[width=\textwidth]{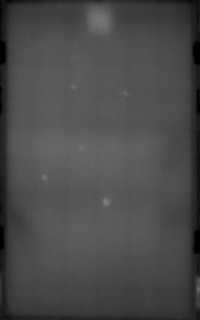}
         \caption{So: Soiling}
     \end{subfigure}
        \caption{Exemplary IR images of the module anomaly classes in our analysis. Temperature ranges from \SI{30}{\celsius} (black) to \SI{60}{\celsius} (white). The figure is adopted from our previous work \cite{Bommes.2021}.}
        \label{fig:dataset_fault_patches}
\end{figure}

In this section we analyze, whether the relative maximum module temperatures (see fig.~\ref{fig:mean_of_max_temps_corrected}) alone are sufficient to accurately identify abnormal modules, and whether they can replace the more complex deep learning-based anomaly classifier from sec.~\ref{sec:mapping_module_defects}. To this end, we manually label all $13463$ modules of plant A as healthy or as abnormal with one out of the ten anomaly classes shown in fig.~\ref{fig:dataset_fault_patches}. For each module a binary anomaly prediction is obtained by comparing its relative maximum module temperature to a specified threshold value. Similarly, for the deep learning classifier we compare the anomaly ratio (see fig.~\ref{fig:module_defects_map}) of the module to a threshold value.

As common in the anomaly detection literature \cite{Golan.2018, Bergman.2020b, Hendrycks.2019b}, we quantify the anomaly detection performance as the area under the receiver operating characteristic (AUROC). This metric is independent of a specific threshold value, and therefore, enables a fair comparison of both classifiers, which depend differently on their threshold values. AUROC is defined as the area under the true positive rate $\textrm{TPR} = \textrm{TP} / (\textrm{TP}+\textrm{FN})$ plotted against the false positive rate $\textrm{FPR} = \textrm{FP} / (\textrm{FP}+\textrm{TN})$ at different threshold values. Here, $\textrm{TP}$ and $\textrm{TN}$ are the numbers of correctly classified abnormal and healthy modules, and $\textrm{FP}$ is the number of healthy modules falsely classified as abnormal and $\textrm{FN}$ the number of abnormal modules falsely classified as healthy.

Tab.~\ref{tab:defect_identification_auroc} reports the resulting AUROC scores for each anomaly class and an overall AUROC score, which considers all anomaly classes. The results indicate that both module temperature distribution and deep learning classifier perform equally and nearly perfect on severe anomalies (Sh, Cs+, Cm+), while they complement each other on the less severe anomaly classes. The deep learning classifier performs better for Pid, Cs and Cm anomalies, which are characterized by low temperature gradients, and are therefore not as accurately identifiable by the temperature distribution. On the contrary, the temperature distribution performs better for D, So, Chs anomalies, which have large temperature gradients and a small spatial extent. The small spatial extent makes detection of these anomalies difficult for a convolutional neural network. Both classifiers perform poorly on homogeneously overheated modules (Mh) because their predictions are based on temperature differences within the image (deep learning classifier) or within the local neighbourhood of modules (temperature distribution). However, using absolute instead of the relative maximum module temperatures allows to accurately identify Mh anomalies (see sec.~\ref{sec:mapping_module_temperatures}).

Summing up, the module temperature distribution can supersede a complex deep learning-based anomaly classifier for the detection of seven out of ten common module anomalies in a PV plant. This is beneficial for practical applications because of the simplicity, higher speed and better interpretability of the temperature distribution. Furthermore, no training is required, which saves the effort of creating a labelled training dataset and circumvents the issue of having to generalize from the training to the test dataset.

\begin{table}[tpb]
\centering
\caption{AUROC scores for the detection of module anomalies in plant A by the module temperature distribution versus a deep learning classifier. Scores of the better classifier and scores above \SI{99}{\percent} are in bold.}
\label{tab:defect_identification_auroc}
\begin{tabular}{l
S[table-format=3.0]
S[table-format=3.2]
S[table-format=2.2]}
\toprule
{Anomaly} & {\# Modules} & \multicolumn{2}{c}{AUROC / \%}\\\cmidrule(lr){3-4}
 & & {Temp. Distribution} & {Deep Learning Clf.}\\\midrule
Mh & 22 & \bfseries 59.04 & 52.73\\
Sh & 32 & \bfseries 99.78 & \bfseries 99.95\\
Pid & 149 & 76.01 & \bfseries 95.92\\
Cm+ & 11 & \bfseries 100.00 & \bfseries 99.61\\
Cs+ & 30 & \bfseries 99.81 & \bfseries 99.90\\
Cm & 420 & 60.71 & \bfseries 86.16\\
Cs & 294 & 61.65 & \bfseries 78.84\\
D & 294 & \bfseries 99.31 & 62.06\\
Chs & 23 & \bfseries 89.64 & 81.53\\
So & 136 & \bfseries 79.71 & 61.29\\
%Shd & 4 & 43.04 & \bfseries 82.34\\
\midrule
Overall & 1411 & 74.31 & \bfseries 78.04\\
\bottomrule
\end{tabular}
\end{table}

%% file: conclusion.tex
\section{Conclusion}
\label{sec:conclusion}

In this work, we developed a method for the automatic extraction and georeferencing of PV modules from aerial IR videos, which can be used for fully automatic PV plant inspection. One possible future improvement of our method is the use of centimeter-accurate RTK-GPS instead of standard GPS, which could reduce the RMSE of module geocoordinates and stabilize the SfM procedure. Similarly, accuracy and stability of the SfM procedure could be improved by using visual videos instead of IR videos, as visual videos provide a higher resolution, wider viewing angle, color information and exhibit lower variation of image intensities \cite{ShihSchon.2001}. However, this requires accurate temporal synchronization and spatial registration of the visual and IR stream, which is a challenging task. Another future direction is the correlation of the obtained temperature distribution with electrical data, such as power and yield, which could provide additional insights into the health state of a PV plant. Finally, our method could be extended for augmented reality applications by rendering a more immersive $3$D model of the plant with overlaid textures, module images and interactive reports for each module.

%Future works could also investigate the impact of changing environmental conditions, such as irradiance, cloud cover, air temperature, and wind speed, on the different anomaly classification methods.

%###########################################################################################

%% file: acknowledgements.tex
\section{Acknowledgements}

This work was financially supported by the State of Bavaria via the project PV-Tera (No. 446521a/20/5) and by BMWi via the project COSIMA (FKZ: 032429A). We sincerely thank the N\nobreakdash-Ergie Nürnberg and PV Service Pro Kollnburg for supporting the project. The authors have declared no conflict of interest.

%% file: appendix.tex
\section{Appendix}

\subsection{Additional Georeferencing Results}
\label{sec:additional_georeferencing_results}

\begin{figure*}[htbp]
     \captionsetup[subfigure]{aboveskip=0pt, belowskip=0pt, justification=centering}
     \centering
     \begin{minipage}{.25\linewidth}
         \begin{subfigure}[t]{\linewidth}
             \centering
             \includegraphics[width=0.9\linewidth, trim=0 1cm 0 1cm, clip]{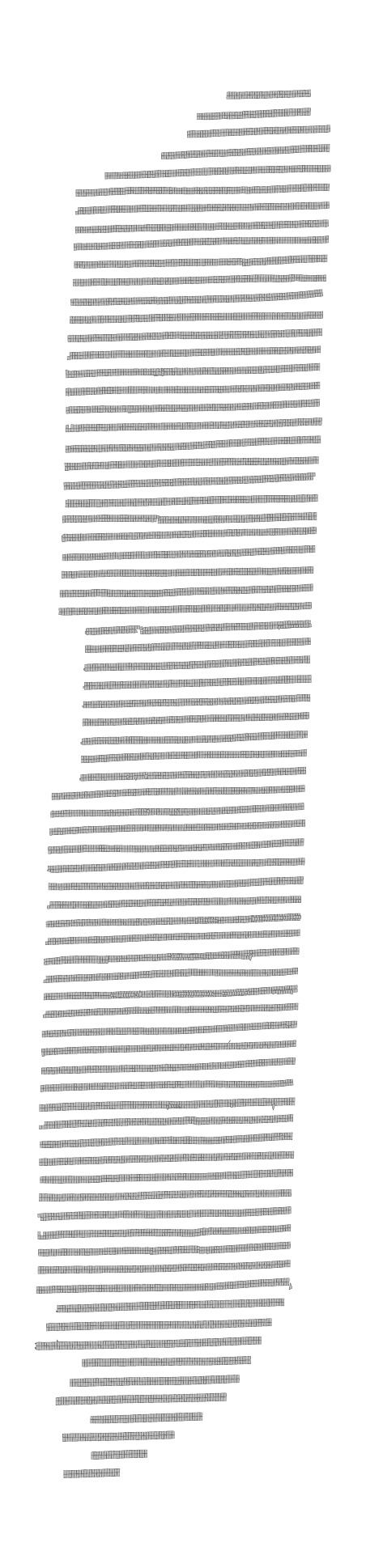}
             \caption{Plant A}
         \end{subfigure}
     \end{minipage}
         \begin{minipage}{.7\linewidth}
         \begin{subfigure}[t]{0.5\linewidth}
             \centering
             \includegraphics[width=\linewidth, trim=1cm 0 1cm 1cm, clip]{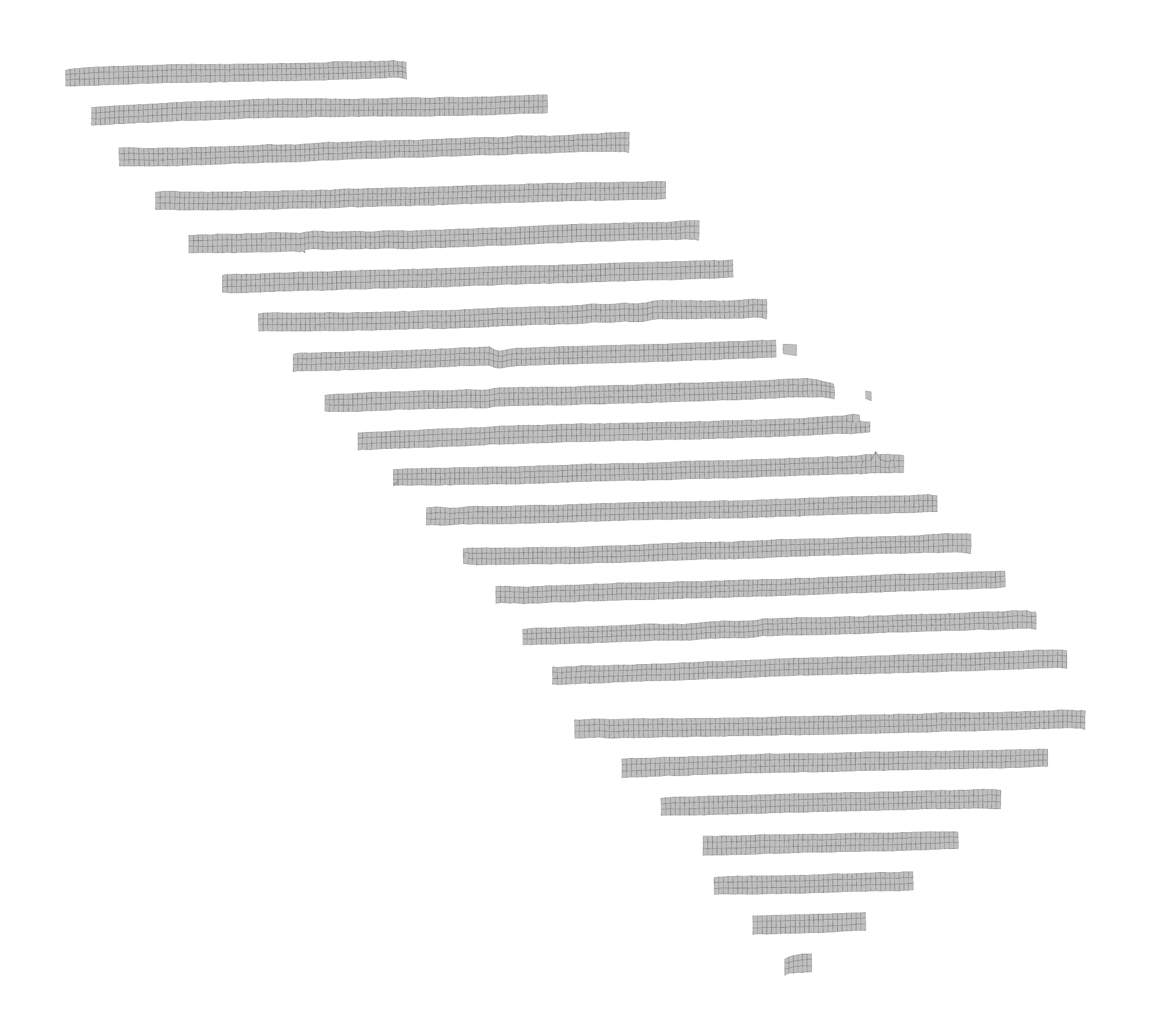}
             \caption{Plant C}
         \end{subfigure}
         \hspace{-2cm}
         \begin{subfigure}[t]{0.5\linewidth}
             \centering
             \includegraphics[width=\linewidth, trim=1cm 0 1cm 1cm, clip]{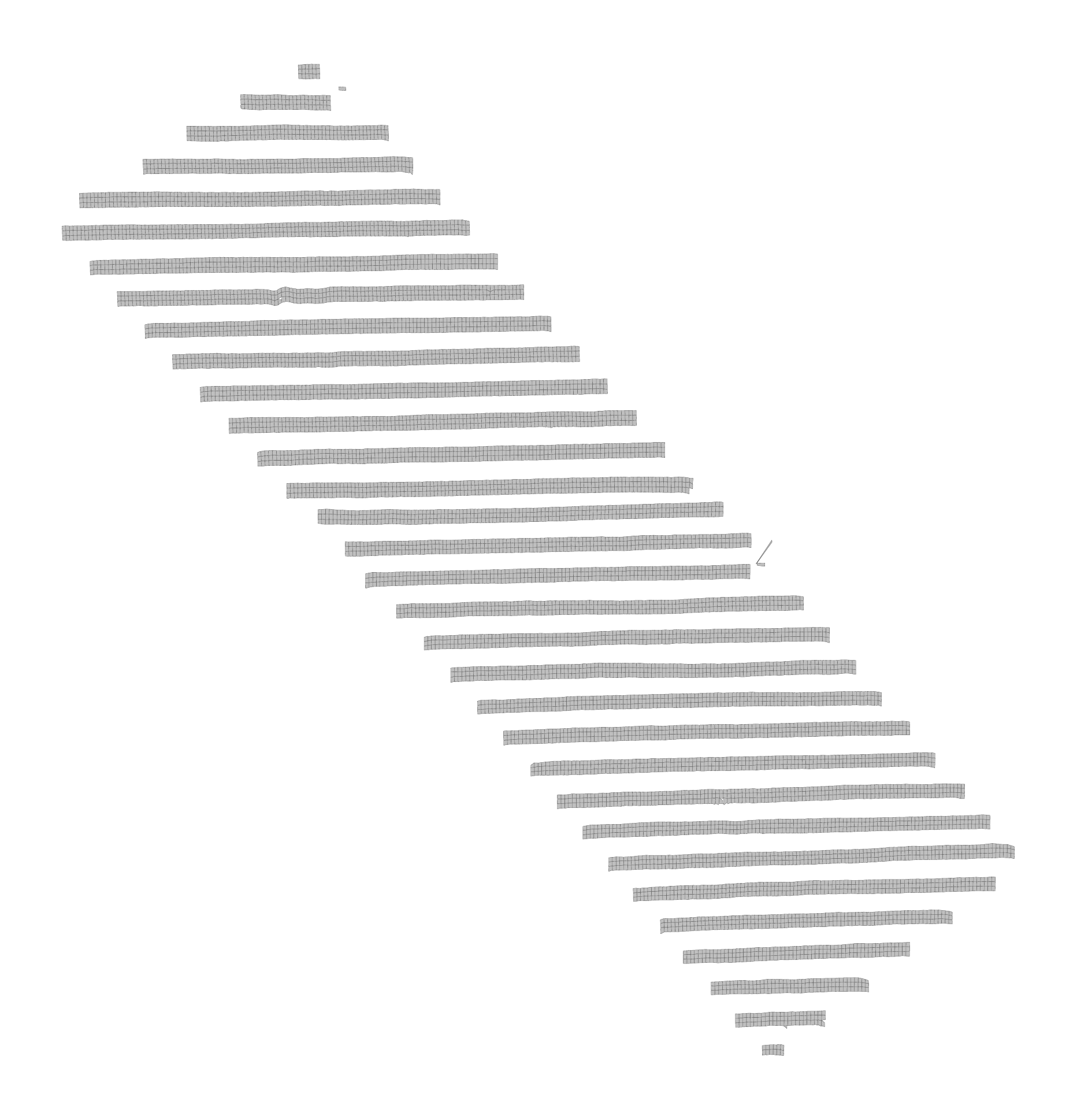}
             \caption{Plant D}
         \end{subfigure} \\
         \hspace*{0.6cm}
         \begin{subfigure}[t]{0.3\linewidth}
             \centering
             \includegraphics[width=\linewidth]{figures/module_layout_plant_B.pdf}
             \caption{Plant B}
         \end{subfigure}
         \hspace{1cm}
         \begin{subfigure}[t]{0.45\linewidth}
             \centering
             \includegraphics[width=\linewidth]{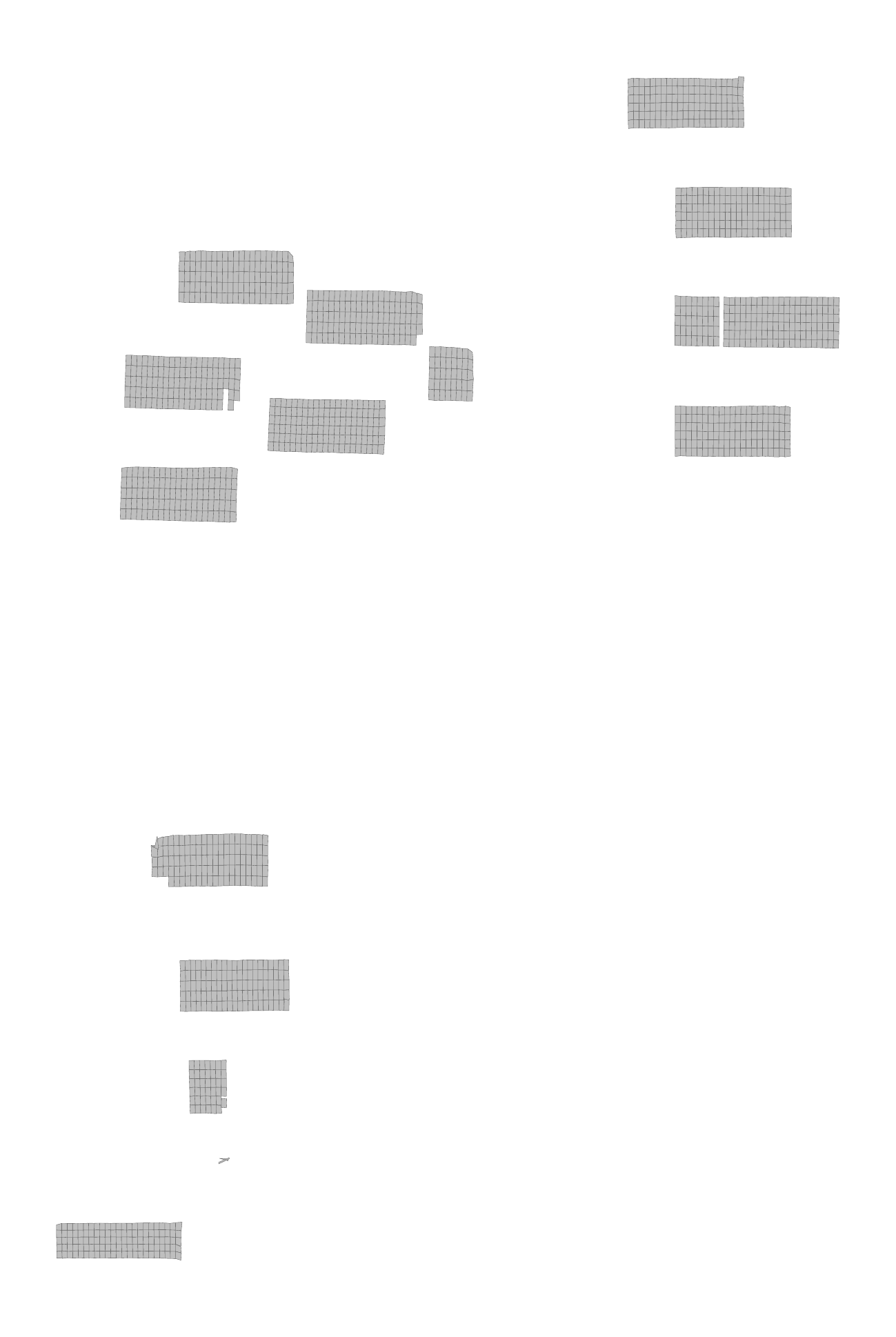}
             \caption{Plant E}
         \end{subfigure}
     \end{minipage}
        \caption{Georeferencing results for all PV plants in our dataset.}
        \label{fig:further_module_layouts}
\end{figure*}